\documentclass[sigconf]{acmart}

\usepackage{url}
\usepackage{graphicx}
\usepackage{appendix}
\usepackage{algorithmic}
\usepackage{subfigure}
\usepackage{amsmath}
\usepackage{xcolor}
\usepackage{booktabs}
\usepackage{multirow}
\usepackage{balance}
\usepackage{listings}
\usepackage{booktabs}
\usepackage{pifont}
\usepackage{enumitem}
\usepackage[ruled,vlined,linesnumbered]{algorithm2e}
\usepackage[inkscapelatex=false]{svg}
\usepackage{bm}
\usepackage{array}
\usepackage{ragged2e}
\usepackage{tabularray}
\usepackage[normalem]{ulem}
\usepackage{enumitem}
\usepackage{blindtext}

\newcommand{\etal}{\textit{et al}. }
\newcommand{\ie}{\textit{i}.\textit{e}.}
\newcommand{\eg}{\textit{e}.\textit{g}.}

\SetAlgoNlRelativeSize{-10}

\begin{document}
\begin{sloppypar}
\definecolor{new_gray}{rgb}{0.8, 0.8, 0.8}

\title{FedConv: A Learning-on-Model Paradigm for Heterogeneous Federated Clients}

\author{Leming Shen\textsuperscript{1},
    Qiang Yang\textsuperscript{1,2},
    Kaiyan Cui\textsuperscript{1,3},
    Yuanqing Zheng\textsuperscript{1}\\
    Xiao-Yong Wei\textsuperscript{4,1},
    Jianwei Liu\textsuperscript{5},
    Jinsong Han\textsuperscript{5}
}

\affiliation{
    \textsuperscript{1}The Hong Kong Polytechnic University,
    \textsuperscript{2}University of Cambridge\\
    \textsuperscript{3}Nanjing University of Posts and Telecommunications,
    \textsuperscript{4}Sichuan University,
    \textsuperscript{5}Zhejiang University
    \country{}
}


\email{
    {leming.shen, qiang.yang, kaiyan.cui}@connect.polyu.hk, csyqzheng@comp.polyu.edu.hk,
}
\email{cswei@scu.edu.cn, {jianweiliu, hanjinsong}@zju.edu.cn}

\renewcommand{\shortauthors}{Leming Shen, Qiang Yang, Kaiyan Cui, Yuanqing Zheng, Xiao-Yong Wei, Jianwei Liu, Jinsong Han}

\begin{abstract}
Federated Learning (FL) facilitates collaborative training of a shared global model without exposing clients' private data. In practical FL systems, clients (\textit{e}.\textit{g}., edge servers, smartphones, and wearables) typically have disparate system resources. Conventional FL, however, adopts a one-size-fits-all solution, where a homogeneous large global model is transmitted to and trained on each client, resulting in an overwhelming workload for less capable clients and starvation for other clients. To address this issue, we propose \textit{FedConv}, a client-friendly FL framework, which minimizes the computation and memory burden on resource-constrained clients by providing heterogeneous customized sub-models. \textit{FedConv} features a novel \textit{learning-on-model} paradigm that learns the parameters of the heterogeneous sub-models via \textit{convolutional compression}. Unlike traditional compression methods, the compressed models in \textit{FedConv} can be directly trained on clients without decompression. To aggregate the heterogeneous sub-models, we propose \textit{transposed convolutional dilation} to convert them back to large models with a unified size while retaining personalized information from clients. The compression and dilation processes, transparent to clients, are optimized on the server leveraging a small public dataset. Extensive experiments on six datasets demonstrate that \textit{FedConv} outperforms state-of-the-art FL systems in terms of model accuracy (by more than 35\% on average), computation and communication overhead (with 33\% and 25\% reduction, respectively).

\end{abstract}

\ccsdesc[500]{Human-centered computing~Ubiquitous and mobile computing}
\ccsdesc[500]{Computing methodologies~Learning paradigms}

\keywords{Federated learning, Model heterogeneity, Model compression}

\acmYear{2024}\copyrightyear{2024}
\acmConference[MOBISYS '24]{The 22nd Annual International Conference on Mobile Systems, Applications and Services}{June 3--7, 2024}{Minato-ku, Tokyo, Japan}
\acmBooktitle{The 22nd Annual International Conference on Mobile Systems, Applications and Services (MOBISYS '24), June 3--7, 2024, Minato-ku, Tokyo, Japan}
\acmDOI{10.1145/3643832.3661880}
\acmISBN{979-8-4007-0581-6/24/06}

\maketitle
\vspace{-5pt}
\section{Introduction}
\label{sec:intro}

Federated Learning (FL) allows mobile devices to collaboratively train a shared global model without exposing their private data \cite{yang2023hierarchical, yang2022federated, cai2022fedadapter, cai2023federated, cai2023efficient}. In each communication round, clients keep their private data locally and only upload their model parameters or gradients to a server after local training. The server then orchestrates model aggregation and updates the global model for the next round \cite{DBLP:conf/aistats/McMahanMRHA17}.

\begin{figure}[t]
\vspace{-5pt}
\setlength{\abovecaptionskip}{2pt}
    \centering
    \includegraphics[width=0.35\textwidth]{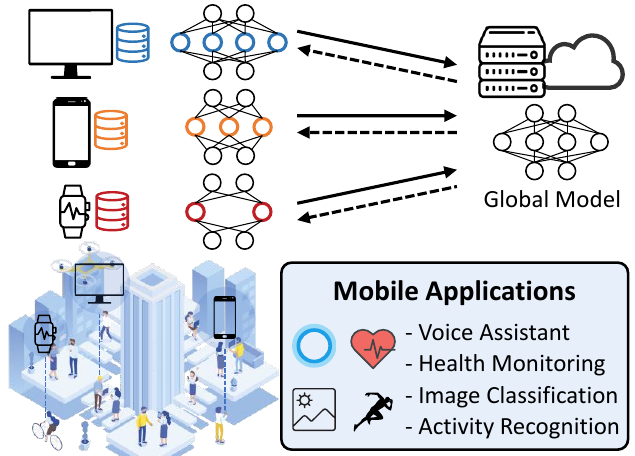}
    \caption{Heterogeneous models in federated learning.}
    \label{fig:model_heterogeneity}
    \vspace{-15pt}
\end{figure}

In real-world deployments, federated clients typically have diverse system resources, calling for heterogeneous models with different sizes \cite{zhang2021edge, jin2022survey, han2024dtmm}. As shown in Fig. \ref{fig:model_heterogeneity}, high-end PCs can support large models, while wearables cannot. Simply assigning the smallest affordable model to all clients results in resource under-utilization and sub-optimal performance.

Previous solutions that generate heterogeneous models mainly include knowledge distillation (KD) \cite{DBLP:conf/nips/LinKSJ20, DBLP:journals/corr/abs-1910-03581}, parameter sharing \cite{DBLP:conf/iclr/Diao0T21}, and parameter pruning \cite{DBLP:conf/mobicom/0005SLPLC21, DBLP:conf/sensys/DengCR0LLZ22}. KD distills the knowledge from heterogeneous client models to a global model for aggregation. Nonetheless, it imposes additional compute overhead on clients \cite{mora2022knowledge} as they must first train on public data and then transfer knowledge via private data. Parameter-sharing strategies distribute different regions of a global model as sub-models to different clients. However, some sub-models can only be trained on a small portion of the dataset. Parameter pruning methods utilize channel or filter level pruning to generate sparse sub-models. However, they suffer from information loss due to the removal of entire channels or filters (\S~\ref{S: limitations}).
Moreover, to determine the pruning structure, clients need to receive the large global model from the server and then perform the pruning operation locally, increasing the overhead of clients.

Ideally, the heterogeneous sub-models should retain the information of the global model in a way that they can be efficiently sent to and trained on resource-constrained clients without any extra overhead.
To this end, we propose \textit{FedConv}, a client-friendly FL framework for heterogeneous models based on a new \textit{learning-on-model} paradigm. \textit{The key insight is that convolution, a technique to extract effective features from data, can also compress large models via various receptive fields while preserving crucial information.} In \textit{FedConv}, the server performs \textit{convolutional compression} on the global model to learn parameters of diverse sub-models according to clients' resource budgets.
Clients directly train on the compressed sub-models as in traditional FL without model decompression.
In model aggregation, the server first uses transposed convolution (TC) to transform heterogeneous client models into large models that have the same size as the global model. Then, the server assigns different learned \textit{weight vectors} to these dilated models and aggregates them.
\textit{FedConv} optimizes the model compression, dilation, and aggregation processes by leveraging a small dataset on the server that can be obtained via crowdsourcing, or voluntarily shared by users without compromising their privacy. Therefore, our system does not incur extra communication or computation overhead for resource-constrained clients.

\begin{figure*}[t]
\vspace{-5pt}
\setlength{\abovecaptionskip}{0pt}
\subfigtopskip=-2pt
\subfigcapskip=-2pt
    \centering
    \subfigure[The parameter sharing scheme]{
        \includegraphics[width=0.225\textwidth]{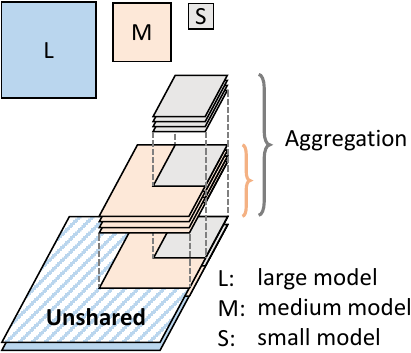}
        \label{fig:heterofl}
    }
    \centering
    \subfigure[Performance difference]{
        \includegraphics[width=0.225\textwidth]{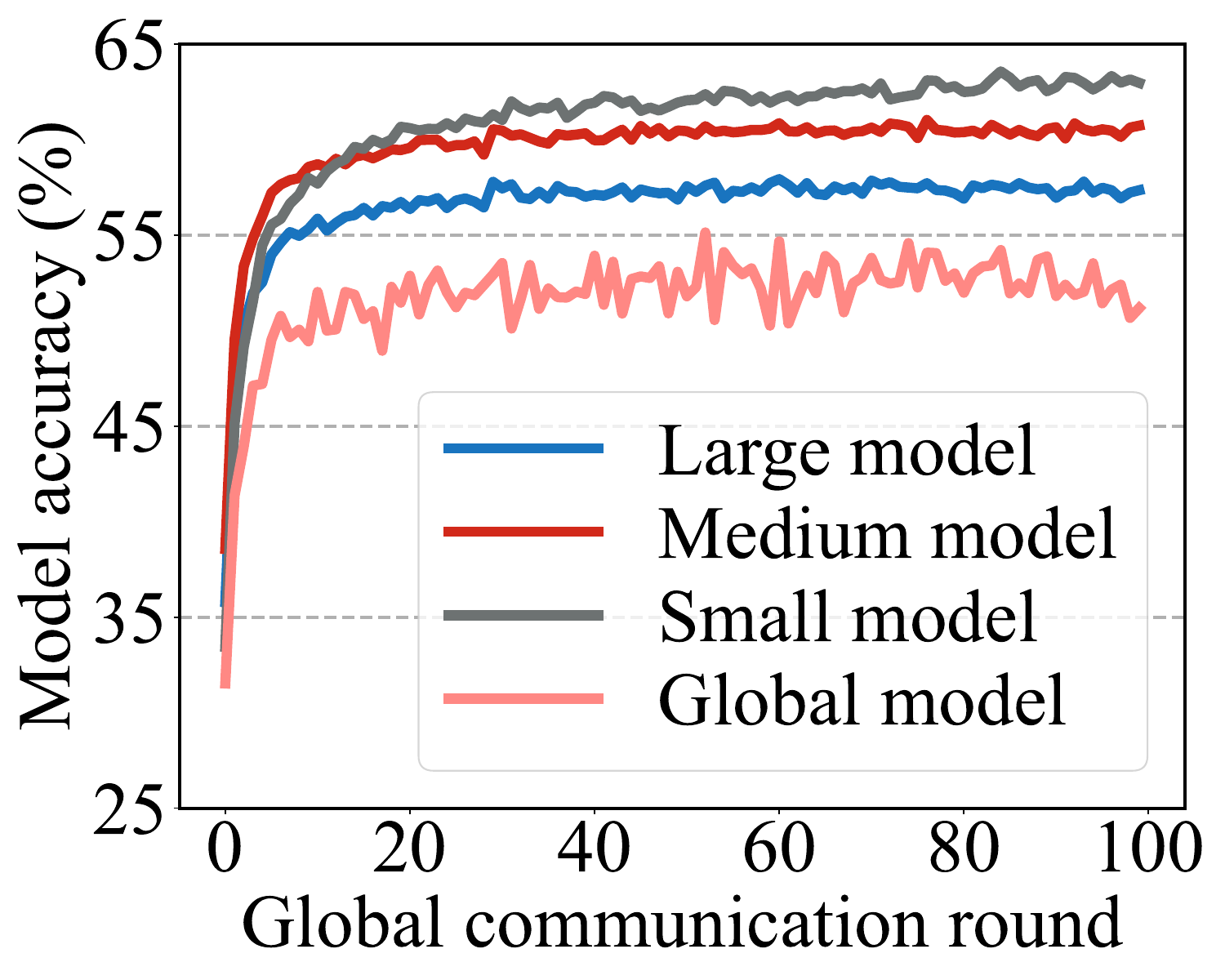}
        \label{fig:performance_difference}
    }
    \centering
    \subfigure[Two pruning schemes]{
    \includegraphics[width=0.225\textwidth]{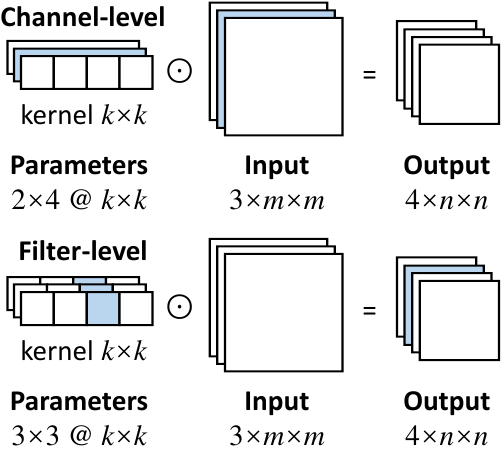}
    \label{fig:pruning_scheme}
    }
    \centering
    \subfigure[Information loss]{
        \includegraphics[width=0.225\textwidth]{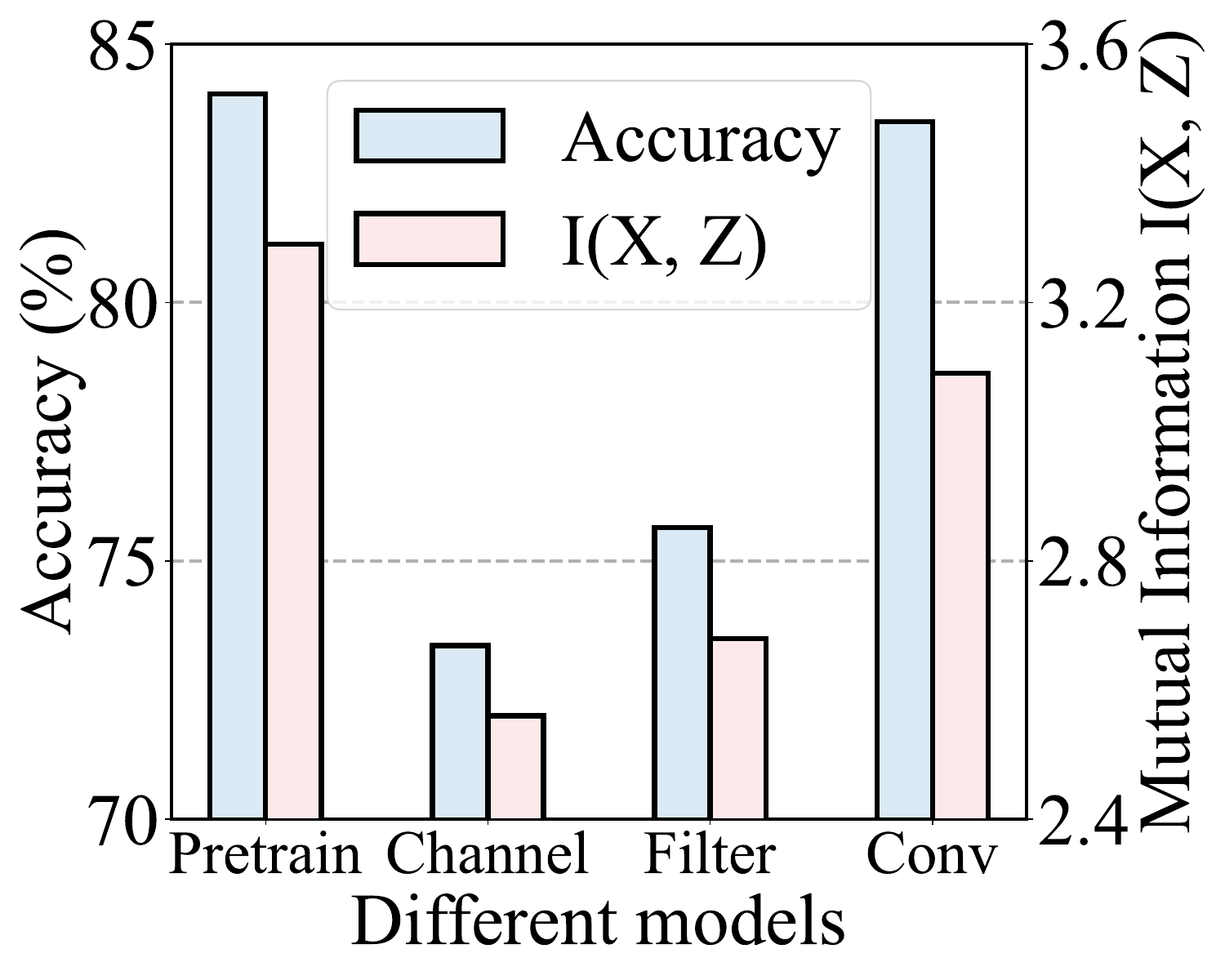}
        \label{fig:information_loss}
    }
    \caption{The parameter sharing and pruning scheme with limitations (the pruned part is colored blue in (c)).}
    \vspace{-8pt}
\end{figure*}

To deliver a practical system, we address three key technical challenges: 1) How to learn the parameters of heterogeneous sub-models via convolution while retaining the global model's prediction capability?
To tackle this problem, we formulate the compression process as a training task. By iteratively fine-tuning the convolution operations, heterogeneous sub-models can be learned effectively and achieve a performance comparable to that of the global model. 2) How to preserve clients' personalized information after converting their models to a unified size for aggregation?
We apply separate TC operations on each client's model parameters and learn a set of dilated models, which inherit their personalized information. We add a residual connection to further enhance the transfer of personalized information from client models to dilated models.
3) How to aggregate these dilated models with imbalanced contributions of heterogeneous federated clients? 
As client models are trained on the non-independent and identically distributed (non-IID) personalized data, 
directly averaging \cite{DBLP:conf/aistats/McMahanMRHA17} these large models would lead to performance degradation. To tackle this issue, we set different learnable \textit{weight vectors} for the dilated models. Through a tuning process, the server can learn the relative importance of each model and orchestrate the final aggregation.

We implement \textit{FedConv}\footnote{The code is available at \href{https://github.com/lemingshen/FedConv}{https://github.com/lemingshen/FedConv}.} based on a user-friendly FL framework (Flower \cite{DBLP:journals/corr/abs-2007-14390}) with two representative FL tasks (image classification and human activity recognition). We evaluate \textit{FedConv} on six public datasets and compare its performance with eight baselines. The experiments show that \textit{FedConv} outperforms the SOTA in terms of inference accuracy (by more than 35\% on average), memory, and communication cost (with 21\% and 25\% reduction, respectively). Besides, \textit{FedConv} substantially reduces the computation overhead for federated clients and saves the total training time.

In summary, we make the following key contributions:
\begin{itemize}[nosep]
    \item To our knowledge, \textit{FedConv} is the first model compression method based on convolution operations. This paradigm can not only compress the global model effectively, but also preserve its crucial information, without imposing extra burden on resource-constrained mobile clients.
    \item \textit{FedConv} handles heterogeneous models with new technologies. Specifically, we propose a \textit{convolutional compression} module to compress the global model and generate heterogeneous sub-models via our \textit{learning-on-model} paradigm. We design a \textit{transposed convolutional dilation} method to obtain models with uniform sizes and use \textit{weighted average aggregation} to balance clients' contributions for final aggregation.
    \item We evaluate \textit{FedConv} based on Flower and conduct comprehensive evaluations with heterogeneous mobile devices. The results demonstrate the superior performance of \textit{FedConv} in terms of both inference accuracy and resource efficiency.
\end{itemize}


\vspace{-5pt}
\section{Motivation}

In this section, we underscore the necessity of model heterogeneity-aware FL systems and analyze SOTA works (parameter sharing and model pruning) to motivate our work. Knowledge distillation-based methods incur heavy overhead on clients (\S~\ref{S:overhead}), which is not suitable for resource-constrained mobile devices.

\vspace{-5pt}
\subsection{Necessity of Heterogeneous Models}
In a conventional FL system, all clients typically share the same model architecture. In practice, however, different clients have diverse computation and communication resources. For example, high-end edge PCs usually have more resources, while low-cost embedded systems have much constrained resources. Therefore, the size of a global model is typically upper-bounded by the clients with the least system resources in conventional FL. Such a one-size-fits-all solution often leads to sub-optimal performance. Moreover, clients with more resources suffer from starvation \cite{shin2022fedbalancer, liu2021fedpa} when waiting for weaker clients in synchronized FL \cite{DBLP:conf/mobicom/LiZZC22, sun2022fedsea, shi2020hysync}. To make full use of more powerful clients while accommodating those with limited resources, it is necessary to develop an FL system that supports heterogeneous models with varied parameter sizes that best fit all clients with diverse resources.

\vspace{-5pt}
\subsection{Limitations of Existing Solutions}
\label{S: limitations}


\textbf{Imbalance problem in parameter sharing.}
HeteroFL \cite{DBLP:conf/iclr/Diao0T21} is a representative parameter sharing scheme where clients share different regions of the global model.
As shown in Fig. \ref{fig:heterofl}, the shared portions (the overlapped part across different sizes of models) are fixed, and the parameters are aggregated only from clients that hold them, missing the information from other clients. To showcase its impact, we train a ResNet18 model \cite{DBLP:conf/cvpr/HeZRS16} on the CIFAR10 dataset \cite{krizhevsky2009learning} with 100 global rounds.
We find that smaller models outperform larger models (Fig. \ref{fig:performance_difference}) due to their exposure to a larger volume of data held by more clients. Besides, the global model exhibits instability and even performs worse than the large model due to unbalanced aggregation. Thus, this scheme will lead to imbalanced performance among clients and unexpected performance degradation \cite{DBLP:journals/corr/abs-2111-14655, alam2022fedrolex}. FedRolex \cite{alam2022fedrolex} proposes a dynamic sharing scheme to tackle the imbalance issue. It enables sub-models to share different parts of the global model's parameters via multiple rolling windows, ensuring that the aggregated parameters are evenly trained on all client-side datasets. However, since different clients contribute distinct parts, the aggregated parameters comprise mixed windows from the diverse sub-models. As a result, the distribution of the global model's parameters is distorted and thus cannot effectively extract useful features from the input data \cite{thibeau2023interpretability}, leading to degraded performance and a longer convergence time.

\begin{figure}[t]
\setlength{\abovecaptionskip}{3pt}
    \centering
    \includegraphics[width=0.47\textwidth]{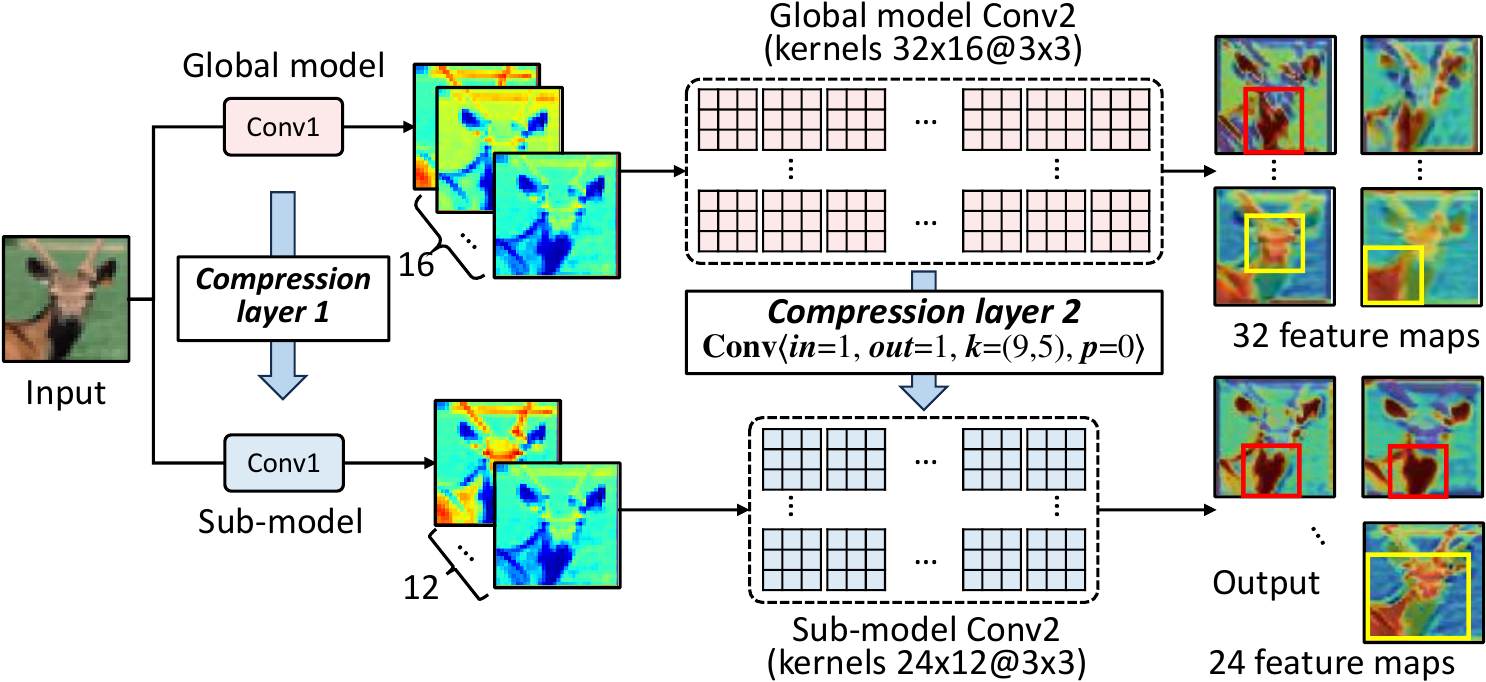}
    \caption{Convolutional compression process.}
    \label{fig:compressed}
    \vspace{-15pt}
\end{figure}

\textbf{Information loss and client workload in model pruning.} 
As shown in Fig. \ref{fig:pruning_scheme}, model pruning can be categorized into channel-level and filter-level pruning. Channel-level pruning removes some input channels from the model parameters, where the corresponding channels of the input data are also excluded from the training process. Filter-level pruning prunes out some output channels (filters), resulting in fewer output feature maps.
Consequently, these two schemes suffer from information loss as they not only discard some input data channels or feature maps but also remove certain weights or connections in model parameters \cite{blakeney2020pruning}.
To study the information loss, we apply channel-level and filter-level pruning to the pre-trained ResNet18 model based on the parameter magnitude ranking. We measure the mutual information (MI) $\mathrm{I}(X,Z)$, which quantifies the amount of information that can be inferred from $X$ after observing $Z$ \cite{10.5555/1146355}. We find that the MI between the parameters of the ResNet18 model and itself is 3.29, whereas the MI between the parameters of the pre-trained model and the channel-pruned model is reduced to 2.56, with an accuracy drop from 84.04\% to 73.36\%. Similarly, the accuracy of the filter-pruned model drops to 75.64\% while MI is 2.68. 
This indicates information loss due to pruning.
Moreover, existing pruning-based methods typically require the server to transmit all the parameters of a global model to clients, and perform model pruning at resource-constrained clients, which incurs high communication and computation overhead for clients.

\vspace{-8pt}
\subsection{Model Compression via Convolution}
Ideally, a compression method should minimize the information loss of model parameters to retain the performance after compression, without posing extra computation or communication burden on resource-constrained clients.
To this end, we propose a novel \textit{convolutional compression} technique that applies convolution operations on the global model parameters to generate the parameters of heterogeneous sub-models while preserving crucial information of the global model (\eg, parameter distributions and patterns).
Our preliminary studies find that by applying refined convolution operations via various receptive fields \cite{DBLP:conf/eccv/ZeilerF14}, the sub-model can inherit spatial and hierarchical parameter patterns from the global model. These receptive fields selectively determine which parameter information should be retained after convolution. Hence, the generated sub-models can also extract valuable features from the input data, similar to the features extracted by the large global model.



Fig.~\ref{fig:compressed} shows the convolution-based compression process. To showcase that a compressed model generated by \textit{convolutional compression} (compression layers) can effectively extract features from the input data, we compress the pre-trained model described in \S~\ref{S: limitations} at a shrinkage ratio of 0.75. We then select the top-4 and top-3 feature maps with the highest importance outputted by a convolutional layer (measured by IG \cite{DBLP:conf/icml/SundararajanTY17}) from the large model and the sub-model, respectively.
As shown in Fig. \ref{fig:compressed}, both the large model and the sub-model can learn and focus on the key features (\eg, the deer's body, head, and horn). Moreover, compared with the large model, the first two feature maps from the sub-model pay more attention (deeper color) to the deer's body and ears. The third feature map can be regarded as a fusion of the last two feature maps from the large model, as it focuses on both the body and head of the deer. This observation indicates that the feature extraction capability of the large model can be effectively preserved and transferred to the sub-model via \textit{convolutional compression}. Besides, the accuracy of the sub-model only decreases by 0.19\% and the mutual information between the parameters of the large model and the sub-model is 3.09 (Fig. \ref{fig:information_loss}), which is much higher than that of the pruned model.
This indicates that our proposed \textit{convolutional compression} method can effectively minimize information loss after model compression. 

\begin{figure*}[t]
\vspace{-8pt}
\setlength{\abovecaptionskip}{2pt}
    \centering
        \includegraphics[width=0.75\textwidth]{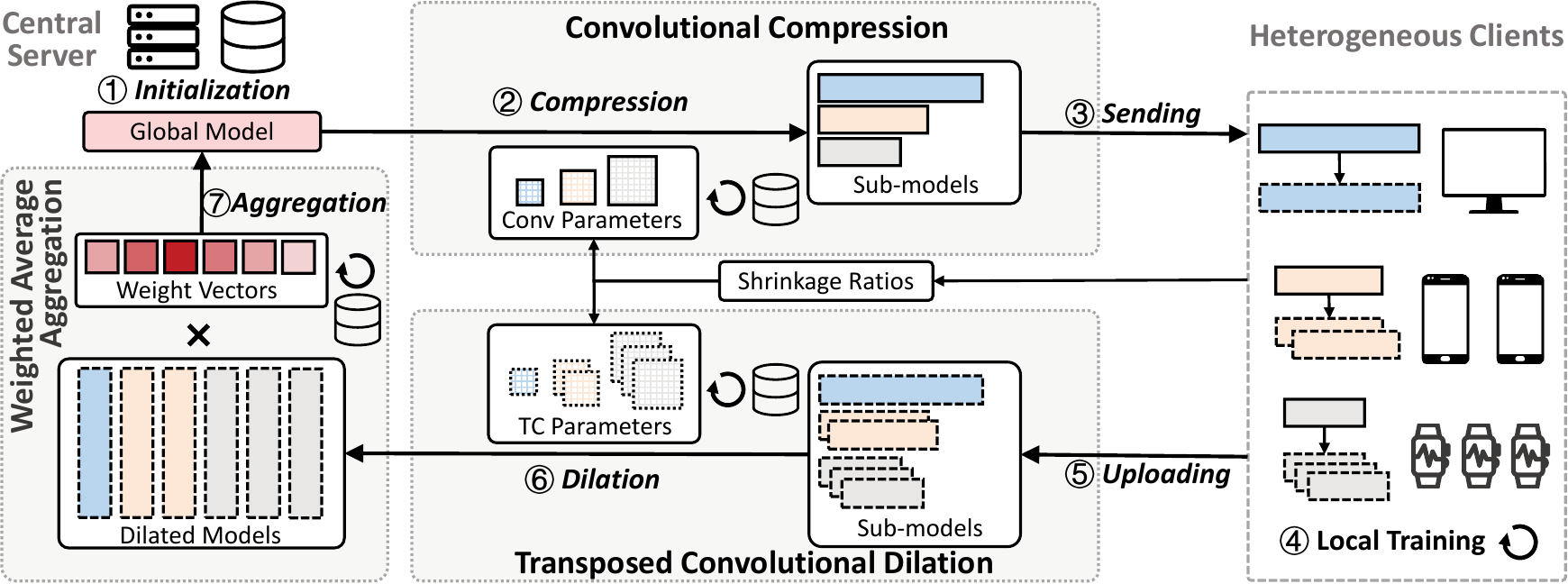}
    \caption{Framework architecture of \textit{FedConv}.}
    \label{fig:architecture}
    \vspace{-12pt}
\end{figure*}

To optimize the compression process, same as existing knowledge distillation-based FL systems \cite{DBLP:journals/corr/abs-1910-03581, DBLP:conf/nips/LinKSJ20} that need server-side data during FL training, we also maintain a small publicly available dataset on the server to fine-tune the compression process (\S~\ref{S: M1}). The server-side data can be collected from public datasets, or crowdsourced by volunteers who are willing to share their data. We note that same as conventional FL schemes, clients do not need to send their data to the server.
By leveraging such server-side data with iterative refining of the compression process, the server can gradually gain a comprehensive global view \cite{DBLP:conf/sensys/DengCR0LLZ22} of the entire FL process and thus transfer more general information from the server to heterogeneous clients \cite{afonin2021towards}.



\section{Framework Overview}

Fig.~\ref{fig:architecture} illustrates the architecture of \textit{FedConv}, consisting of three main modules: \textit{convolutional compression} (\S~\ref{S: M1}), \textit{transposed convolutional dilation} (\S~\ref{S: M2}), and \textit{weighted average aggregation} (\S~\ref{S: M3}).


The server first initializes a global model with an estimated memory requirement and records a set of shrinkage ratios (SR) reported by each client based on their resource profiles (\ding{172}). In the first communication round, the server pre-trains the global model for several epochs with a server-side dataset to gain a better global view of the data distribution \cite{DBLP:conf/sensys/DengCR0LLZ22}. Then, based on the SRs, a set of fine-tuned \textit{convolution parameters} are used to compress the global model with the \textit{convolutional compression} module, and generate heterogeneous sub-models (\ding{173}). Afterwards, the server sends the heterogeneous sub-models to federated clients (\ding{174}). Clients then perform several epochs of local training with their private training dataset to fine-tune the received sub-models (\ding{175}), and then upload the updated parameters to the server (\ding{176}). After that, the server performs the \textit{transposed convolutional dilation}, where different \textit{transposed convolution parameters} are used to dilate the sub-models to a set of large models that have the same size as the global model (\ding{177}). Finally, the server applies the \textit{weighted average aggregation} to aggregate the dilated models with the learned weights(\ding{178}).

In \textit{FedConv}, the compression and dilation operations are transparent to clients and performed by the powerful server, which can be seamlessly integrated into conventional FL systems where clients only need to perform local training.

\vspace{-5pt}
\section{Framework Design}


\subsection{Convolutional Compression}
\label{S: M1}

The \textit{convolutional compression} module leverages a set of convolutional layers (termed as \textit{compression layers}) to compress the global model and generate heterogeneous sub-models. As shown in Fig. \ref{fig:compressed}, after feeding the global model parameters to the \textit{compression layers}, the compressed parameters of the sub-model become smaller and output fewer feature maps. 
We use the server-side data to iteratively optimize the \textit{convolution parameters} (\ie, the parameters of \textit{compression layers}) until the sub-models can achieve comparable performance to the global model, as they inherit the parameter information from the global model with a comprehensive perspective. Thus, they are able to extract general features and can be further updated by clients to fit their local data for personalization.

\textbf{Convolution configurations.} To determine the sizes of the sub-models, clients first specify their shrinkage ratios (SRs). Specifically, the server first broadcasts the size of the global model to all the clients. Then, each client will determine an appropriate SR for the corresponding sub-model to meet its own
computing resource budget\footnote{The resource profiling process can be performed by existing tools, \eg, nn-Meter \cite{DBLP:conf/mobisys/ZhangHWZCYL21}.} (\eg, GPU memory, network bandwidth). Subsequently, the SRs are transmitted back to the server. We note that same as conventional FL schemes, no client-side sensor data needs to be transferred to the server. Accordingly, the server determines the corresponding configurations (\ie, input channel \bm{$in$}, output channel \bm{$out$}, kernel size (\bm{$k_1$}, \bm{$k_2$}), stride \bm{$s$}, and padding \bm{$p$}) of the \textit{compression layers} so that the sizes of generated sub-models match with the expected SRs.
Let's take convolution layers as an example. As shown in Fig. \ref{fig:compressed}, a convolutional layer in the global model has 16 input and 32 output channels with a kernel size of $(3,3)$. Regarding each element in the kernel as a single unit, we can reshape its parameter matrix from $(32,16,3,3)$ to $9\times(1,32,16)$. Suppose the SR is 0.75, the shape of the parameter matrix becomes $9\times(1,24,12)$ after compression.
Therefore, we use nine separate 2D convolutional layers (\ie, \textit{compression layers}) to compress the reshaped matrix. The configuration\footnote{By default, we set the stride and padding as one and zero, respectively. We will vary their values and the kernel size to explore the impact in \S~\ref{S:varying_hyper}} of each \textit{compression layer} is Conv$\langle\bm{in}$=1, $\bm{out}$=1, $\bm{k}$=(9, 5), $\bm{s}$=1, $\bm{p}$=0$\rangle$. This convolution-based process can also be applied to compress other types of layers by properly adjusting the configurations. Note that the input channels of the first layer will not be compressed, ensuring that all channels of the raw data can be fed into the sub-model. Similarly, the output channels of the last layer are also uncompressed, ensuring that the sub-models and the global model have the same prediction task.

\textbf{Convolution parameter fine-tuning.} Next, we need to fine-tune the \textit{convolution parameters} so that the generated sub-models inherit the parameter information from the global model and achieve comparable performance. We use the server-side data to iteratively adapt the \textit{convolution parameters} by minimizing the loss between the ground truth and the prediction result of the compressed model:
\begin{equation}
\small
\setlength{\abovedisplayskip}{2pt}
\setlength{\belowdisplayskip}{2pt}
\begin{split}
    & \min_{\boldsymbol{w}_{Conv,l}}{\sum_{x}{\mathcal{L}(f(x;\boldsymbol{W}_{G,l}\odot \boldsymbol{w}_{Conv,l}), y)}}, \\
    & s.t.\;\forall l\in \{1, 2, \cdots, L\}, \forall (x,y)\in\mathcal{D}.
\end{split}
\end{equation}
where $\mathcal{L}$ is the Cross-entropy loss \cite{enwiki:1170369413} and $f(\cdot)$ is the forward function of the compressed model. $(x, y)$ is the data and the corresponding label in the server-side data $\mathcal{D}$. $\boldsymbol{W}_{G,l}$ is the parameters of the $l$-th layer in the global model. $\boldsymbol{w}_{Conv,l}$ is the \textit{convolution parameters}. $\odot$ is the convolution operation. To fine-tune the \textit{convolution parameters}, the compressed sub-model ($\boldsymbol{W}_{G,l}\odot \boldsymbol{w}_{Conv,l}$) is first evaluated on the server-side data. By back-propagating the calculated loss, the \textit{convolution parameters} $\boldsymbol{w}_{Conv,l}$ is updated while others (\ie, the parameters of the global model and the sub-model) are frozen.

\begin{figure}[t]
\vspace{-8pt}
\setlength{\abovecaptionskip}{-3pt}
\subfigtopskip=-3pt
\subfigcapskip=-3pt
    \subfigure[The convolution process]{
        \label{fig:convolution_process}
        \includegraphics[width=0.2\textwidth]{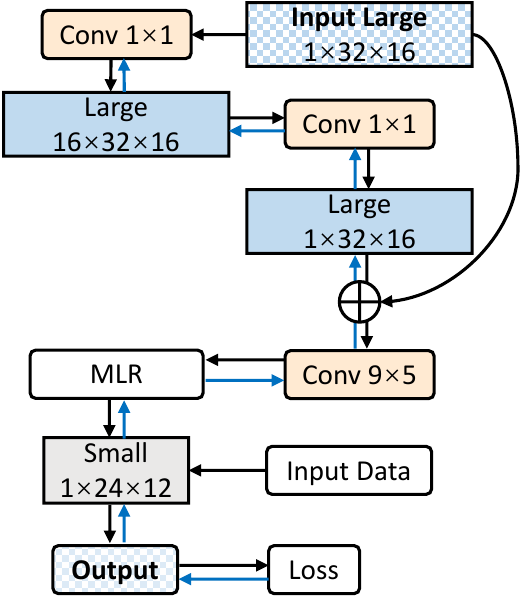}
    }
    \centering
    \subfigure[The TC process]{
        \label{fig:transposed_process}
        \includegraphics[width=0.2\textwidth]{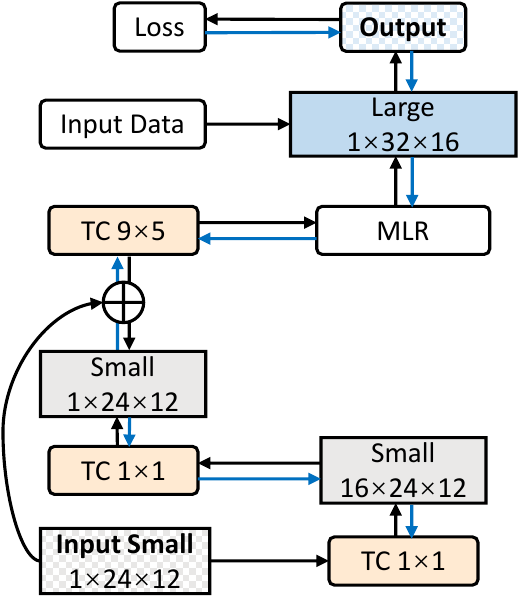}
    }
    \caption{An example of the convolution/TC process (black arrow: forward, blue arrow: backward, blue box: larger parameters, grey box: smaller parameters, orange: Conv/TC).}
    \vspace{-15pt}
\end{figure}

\textbf{Remarks.} In the model compression process, the server applies \textit{compression layers} on the global model parameters and iteratively fine-tunes the \textit{convolution parameters} to learn heterogeneous sub-models, aiming to preserve crucial parameter information and prediction capability from the global model. Such a \textit{learning-on-model} method fundamentally differs from the traditional \textit{learning-on-data} paradigm. 
Specifically, the \textit{learning-on-data} method takes raw data as input and trains a model to extract features, while our \textit{learning-on-model} paradigm takes model parameters as input and uses \textit{compression layers} to generate sub-model parameters. The parameters of \textit{compression layers} are fine-tuned by minimizing the loss between the sub-model outputs and the ground-truth labels.

\textbf{Challenges.} Nevertheless, several practical challenges emerge during this compression process. We use a pre-trained model on the MNIST \cite{DBLP:journals/spm/Deng12} dataset (with an accuracy of 99.04\%) as an example to show how we address these challenges and the progress we make.


\textit{(1) Information loss.} 
After fine-tuning the \textit{convolution parameters} as aforementioned, we find that the amount of the parameter information in the sub-model inferred from the global model is still low (the mutual information between the parameters of the global model and the sub-model is only 0.84). This can be attributed to the limited capability of simple \textit{compression layers} to capture more fine-grained parameter information effectively, leading to a lower accuracy of the sub-model (90.2\%).
To address this issue, we add two $1 \times 1$ convolutional layers with biases before compression (Fig. \ref{fig:convolution_process}). Intuitively, the first Conv $1\times 1$ increases the number of output channels to 16, capturing more diverse and complex parameter information in the global model. The second Conv $1\times 1$ decreases the channel number back to one, fusing information from different channels and producing a comprehensive parameter representation. In addition, we add a residual connection between the global model parameters and the output of the second Conv $1 \times 1$, facilitating the transfer of parameter information from the global model to sub-models through convolution operations. With these designs, the accuracy of the sub-models increases to 93.15\%, indicating that the information loss is effectively mitigated.


\begin{figure}[t]
\vspace{-8pt}
\setlength{\abovecaptionskip}{-3pt}
 \subfigtopskip=-3pt
\subfigcapskip=-3pt
    \centering
    \subfigure[Parameter distribution]{
    \label{fig:activation_function_effect}
        \includegraphics[width=0.22\textwidth]{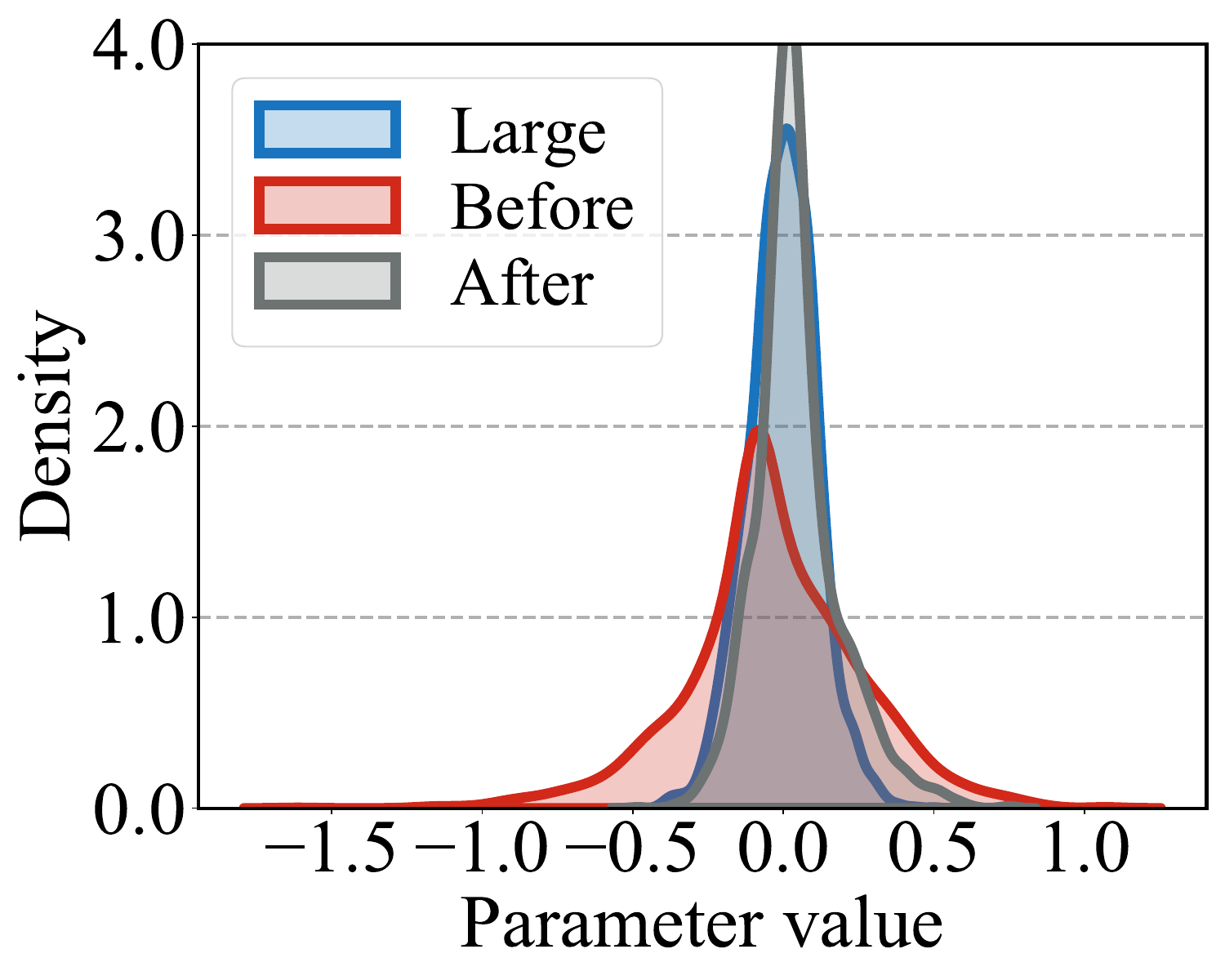}
    }
    \centering
    \subfigure[Modified activation function]{
        \label{fig:modified_activation}
        \includegraphics[width=0.22\textwidth]{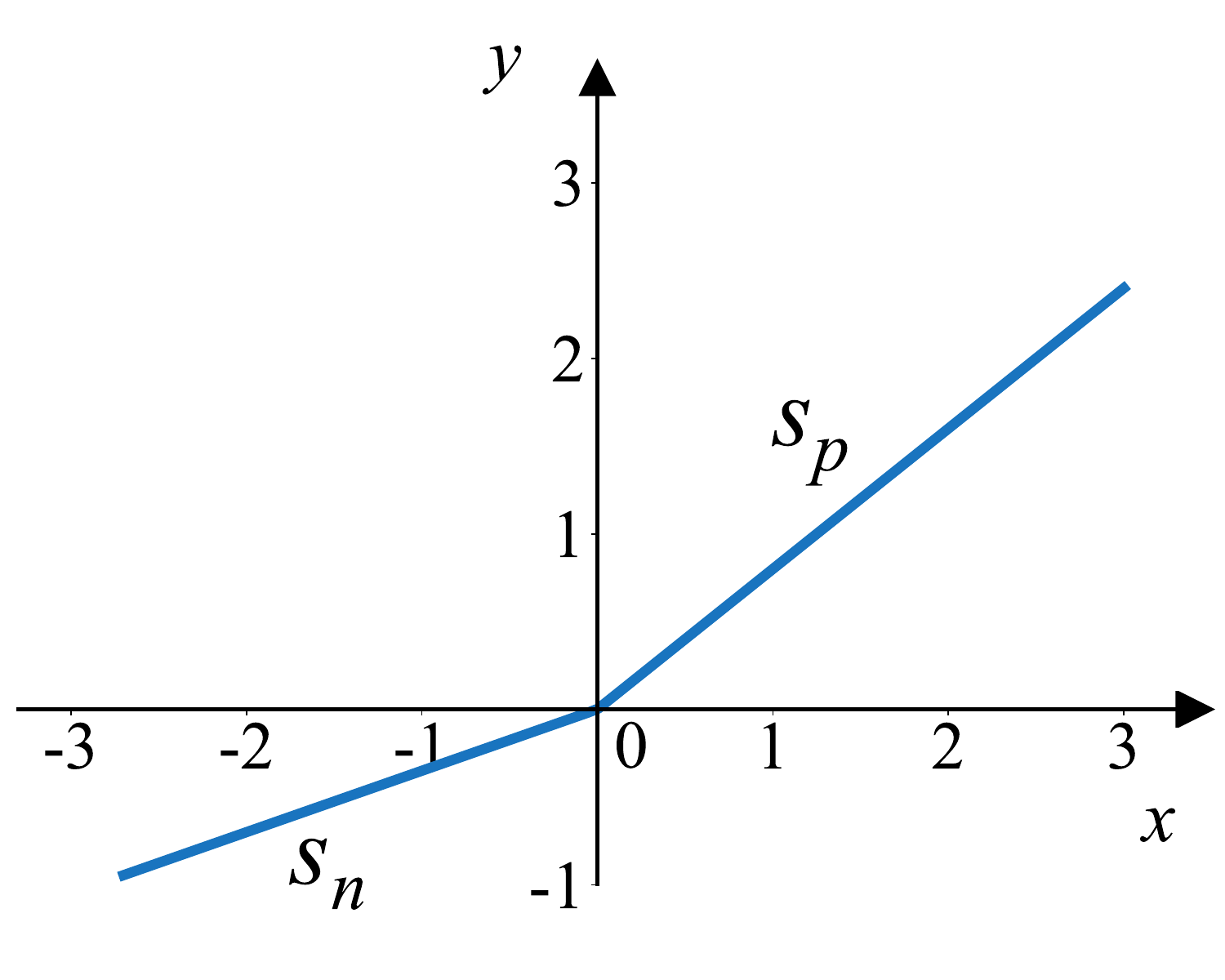}
    }
    \caption{(a) After using the MLR, the distribution of the parameters in the sub-model becomes similar to the large global model (blue curve); (b) The MLR function.}
    \label{fig:convolution_contrast}
    \vspace{-5pt}
\end{figure}

\begin{figure}[t]
\vspace{-5pt}
\setlength{\abovecaptionskip}{-3pt}
\subfigtopskip=-3pt
\subfigcapskip=-3pt
    \centering
    \subfigure[Compressed model's accuracy]{
        \label{fig:compression_challenge}
        \includegraphics[width=0.22\textwidth]{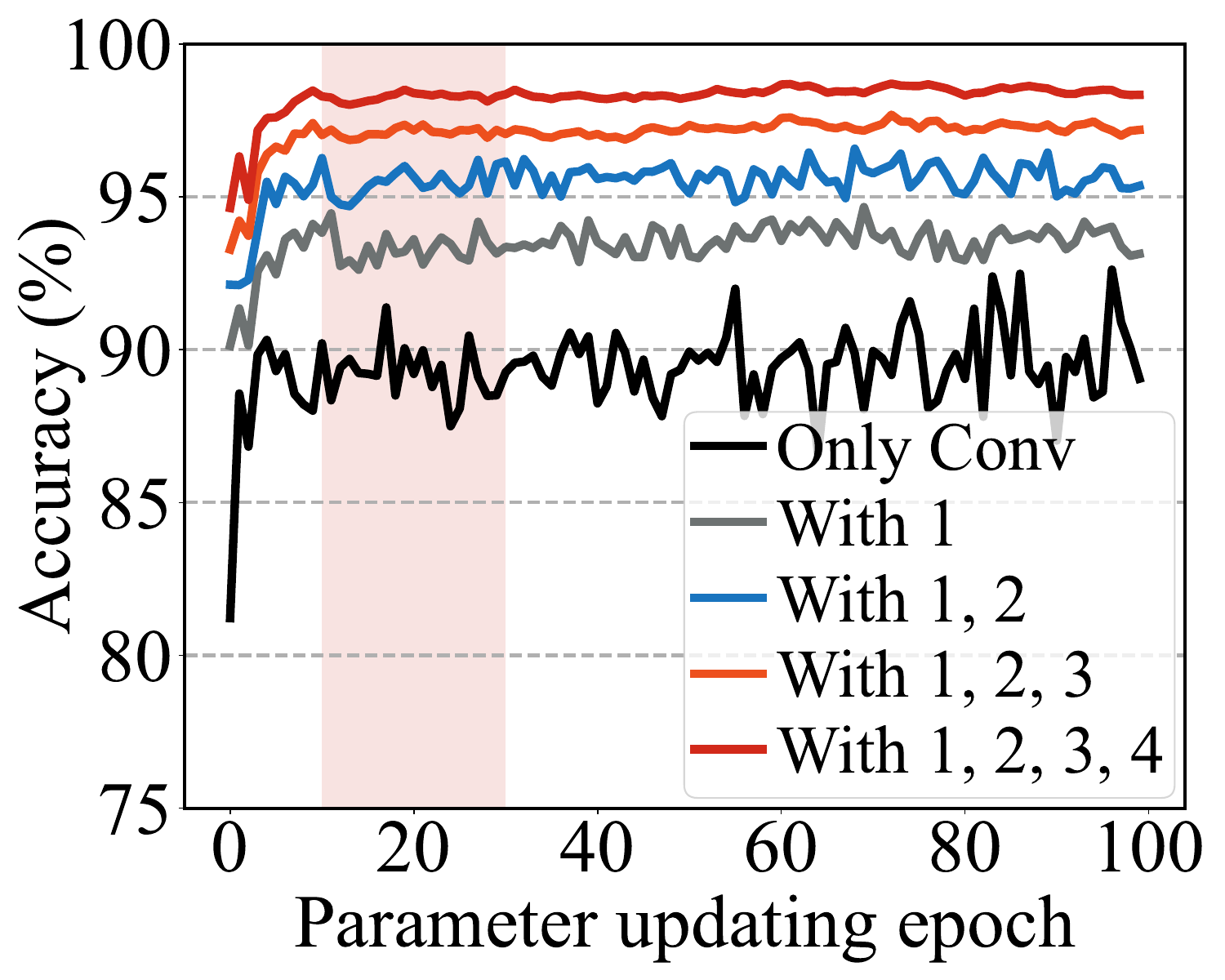}
    }
    \centering
    \subfigure[Different parameter distribution]{
        \label{fig:reconstructed_distribution}
        \includegraphics[width=0.22\textwidth]{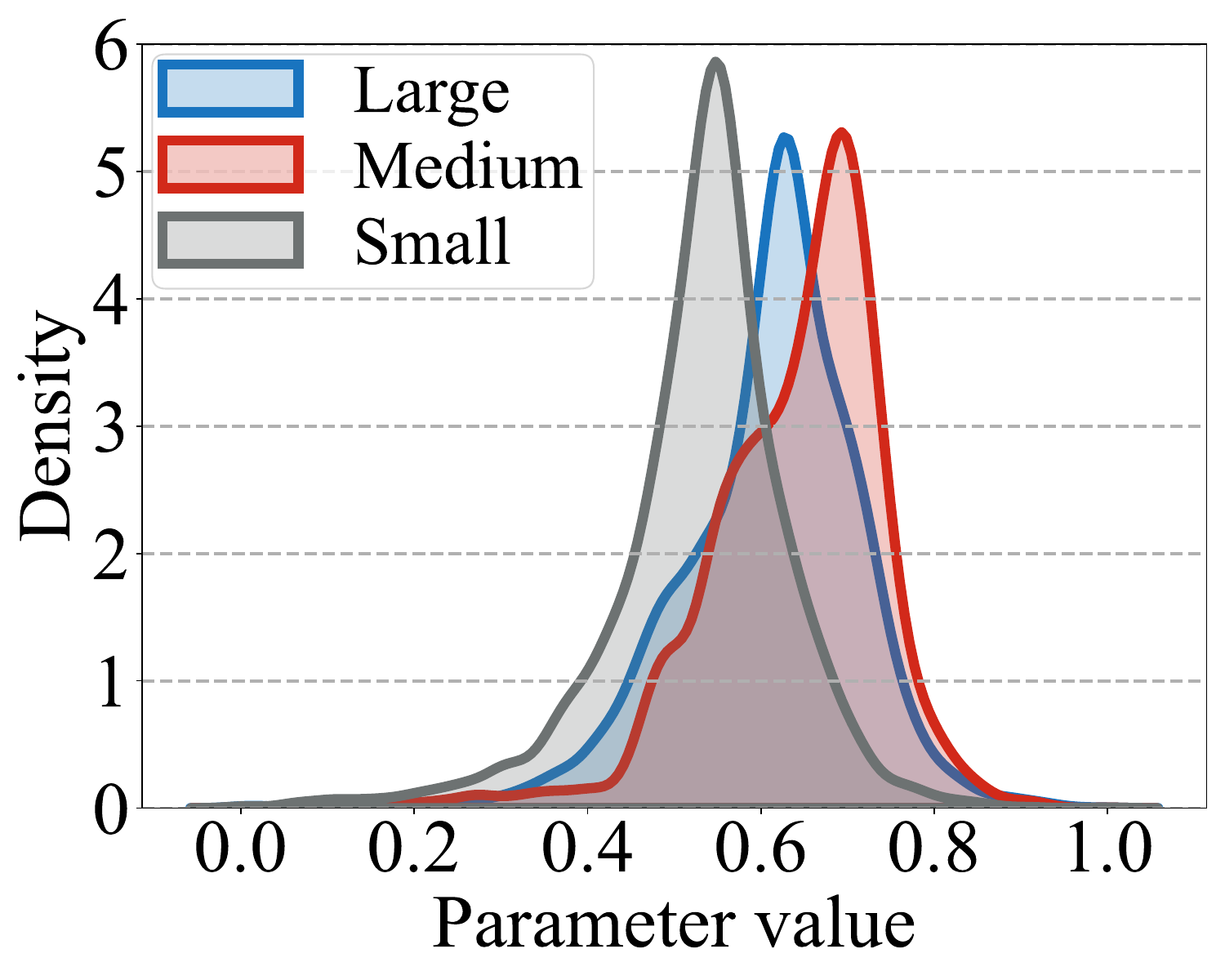}
    }
    \caption{(a) Accuracy of the compressed model with: 1) two Conv $1 \times 1$, 2) the MLR function, 3) weight normalization, 4) learning rate scheduler; (b) The parameter distribution of the dilated models from different sizes of client models.}
    \vspace{-15pt}
\end{figure}

\textit{(2) Imbalanced parameter distribution.} 
As shown in Fig. \ref{fig:activation_function_effect}, although the distributions of the parameters in sub-models and the global model are similar, the parameters in sub-model skew towards negative values, leading to numerical instability, slow convergence, and unexpected performance degradation \cite{DBLP:journals/jmlr/GlorotB10}. Therefore, we adopt a modified Leaky ReLU (MLR) activation function (Fig. \ref{fig:modified_activation}) to rectify the negative value, where $s_n$ and $s_p$ are slopes for negative and positive values, respectively. With a small $s_n$, the MLR can suppress the negative parameters but not entirely eliminate them, thereby preserving potential information embedded in negative parameters. After applying the MLR, the sub-model parameters exhibit a similar distribution pattern and value range to the global model (Fig. \ref{fig:activation_function_effect}). The accuracy of the sub-model further increases to 95.06\%.

\textit{(3) Performance fluctuation.} During the fine-tuning process, we observe significant performance fluctuation of the sub-model. This is because in \textit{learning-on-data} methods, model parameters are directly updated during training. However, in our \textit{learning-on-model} method, only the \textit{convolution parameters} are updated, which subsequently generate sub-model parameters via the convolution process. As a result, the performance of the sub-model exhibits much higher sensitivity to the changes in \textit{convolution parameters}. To address this issue, we apply weight normalization \cite{salimans2016weight} on the \textit{convolution parameters} to decouple their magnitude and direction during updating, which stabilizes the convergence in a fine-grained way. Moreover, we apply a cosine annealing learning rate scheduler \cite{DBLP:conf/iclr/HuangLP0HW17} that dynamically varies the learning rate to avoid local optima and enables faster convergence \cite{DBLP:conf/nips/KrizhevskySH12}. The learning rate undergoes a cosine function decay as the epoch progresses:
\begin{equation}
\small
\setlength{\abovedisplayskip}{2pt}
\setlength{\belowdisplayskip}{2pt}
    lr=lr_{min} + 0.5(lr_{max} - lr_{min})(1 + \cos{(e / T_{max} \cdot \pi)})
    \label{equ:learning_rate_scheduler}
\end{equation}
where $e$ is the current epoch index, $lr_{min}$ and $lr_{max}$ are the lower and upper bound of the learning rate, and $T_{max}$ is the maximum number of iterations before the $lr$ restarts to $lr_{min}$. As shown in Fig. \ref{fig:compression_challenge}, after applying weight normalization and the learning rate scheduler, the accuracy of the sub-model improves to 96.8\% and 98.59\%, respectively. Meanwhile, the performance becomes more stable, and the sub-model converges after around 20 epochs.

After fine-tuning the \textit{convolution parameters}, the compressed models will have comparable performance to the global model. The server then sends the compressed parameters to the corresponding clients for local training. The fine-tuned \textit{convolution parameters} will be kept on the server and updated in the next communication round. Note that this process is performed completely on the server without imposing any extra computation or communication burden on clients. The detailed process is shown in Algorithm~\ref{alg:algorithm2}.

\vspace{-5pt}
\subsection{Transposed Convolutional Dilation}
\label{S: M2}
Upon receiving the updated sub-models from clients, we need to rescale the heterogeneous client models to a unified size for further aggregation. Although knowledge distillation-based methods \cite{DBLP:journals/corr/abs-1910-03581, DBLP:conf/nips/LinKSJ20} are promising, they impose significant computational and communication overhead on clients (\S~\ref{sec:intro}). Instead, we use transposed convolution (TC) layers on the server side, a reverse operation to the \textit{convolution compression}.
In contrast, we apply different \textit{TC layers} to each of the received client models, as they are trained on non-IID data with different sensing heterogeneity and thereby inherently carry diverse personalized information. Then, by meticulously fine-tuning the \textit{TC parameters} (\ie, parameters of the \textit{TC layers}), the personalized information embedded in each client model's parameters will be preserved and transferred to the dilated models for subsequent aggregation.


\textbf{TC configurations.} To transform the heterogeneous models from different clients to a unified size, it is important to ensure that the configurations of each \textit{TC layer} for dilation are identical to the corresponding \textit{compression layer}. For instance, as illustrated in Fig. \ref{fig:transposed_process}, a convolutional layer in a client model has 12 input and 24 output channels with a kernel size of $(3,3)$. With the SR as 0.75, the configurations of the \textit{TC layer} should be TC$\langle\bm{in}$=1, $\bm{out}$=1, $\bm{k}$=(9, 5), $\bm{s}$=1, $\bm{p}$=0$\rangle$. Similarly, this process can also be employed to dilate other kinds of network layers. Note that the input channel number of the first layer and the output channel number of the last layer in all client models are also unchanged.

\begin{algorithm}[h]
\small
    \SetAlgoLined
    \SetKwInOut{Input}{Input}\SetKwInOut{Output}{Output}
    \Input{Global round $r$, pre-training epochs $e_p$, \textit{convolution parameters} updating epochs $e_c$, device type number $n$, SR list $\{SR_1, SR_2, \cdots, SR_n\}$, $T_{max}$, $lr_{max}$, $lr_{min}$}
    \Output{Compressed parameters $\{P_1, P_2, \cdots, P_n\}$}

    /* Configuration initialization */\;
    \eIf{$r$ == 1}{
        Initialize the global model as $\boldsymbol{w}(r)$\;
        Pre-train $\boldsymbol{w}(r)$ for $e_p$ epochs on $\mathcal{D}$\;
        \For{device\_type  $i \in \{1, 2, \cdots, n\}$ \textbf{parallel}}{
        $current\_shrinkage\_ratio \leftarrow SR_i$ \;
            $Conv_i(r)\leftarrow \mathbf{Initialize\_Conv}(\boldsymbol{w}(r), SR_i)$\;
        }
    }
    {
        $Conv_i(r)\leftarrow Conv_i(r-1)$\;
    }

    /* Convolution parameters fine-tuning */\;
    \For{device\_type  $i \in \{1, 2, \cdots, n\}$ \textbf{parallel}}{
        \For{e $\in \{1, 2, \cdots, e_c\}$}{
            $lr=lr_{min} + 0.5(lr_{max} - lr_{min})(1 + \cos{(e / T_{max} \cdot \pi)})$\;
            \For{\textbf{each} $(x, y)$ \textbf{in} $\mathcal{D}$}{
                $P_i \leftarrow\mathbf{MLR}(\boldsymbol{w}(r) \odot Conv_i(r))$\;
                $output \leftarrow f(P_i; x)$\;
                $loss \leftarrow \mathbf{Loss\_fn}(output, y)$\;
                Back-propagate gradient $g_i$ to $Conv_i(r)$\;
                $Conv_i(r)\leftarrow Conv_i(r) - lr\cdot g_i$\;
            }
        }
    }
    Send $P_i$ to the corresponding client\;
    \caption{\textbf{Convolutional compression}}
    \label{alg:algorithm2}
\end{algorithm}

\textbf{TC parameter fine-tuning.}
To fine-tune the \textit{TC parameters}, we also set them as learnable variables and minimize the loss between the ground truth and the prediction result of the dilated large model:
\begin{equation}
\small
\setlength{\abovedisplayskip}{2pt}
\setlength{\belowdisplayskip}{2pt}
\begin{split}
    & \min_{\boldsymbol{w}_{TC,l}}{\sum_{x}{\mathcal{L}(F(x;\boldsymbol{W}_{C,l}\circledcirc \boldsymbol{w}_{TC,l}), y)}}, \\
    & s.t.\;\forall l\in \{1, 2, \cdots, L\}, \forall (x,y)\in\mathcal{D}.
\end{split}
\end{equation}
where $F(\cdot)$ is the forward function of the dilated large model, $\boldsymbol{W}_{C,l}$ is the parameters of the client model, $\boldsymbol{w}_{TC,l}$ denotes the \textit{TC parameters}, and $\circledcirc$ represents the TC operation. To further enhance the integration of personalized information into the dilated models, we also add two TC $1 \times 1$ layers with a residual connection before the dilation process (detailed in Fig. \ref{fig:transposed_process}).

\begin{table*}[htbp]
\vspace{-5mm}
\setlength{\abovecaptionskip}{0pt}
\caption{The hardware configuration of heterogeneous devices in a real-world experiment.}
    \label{tb:hardware_configuration}
\footnotesize
\centering
\begin{tabular}{ccccccccc} 
\toprule[1pt]

Type                   & Device Name            & Number & CPU                             & RAM   & GPU                             & GDDR     & Network  & SR    \\ 
\hline
Server                 & ASUS W790-ACE Server   & 1      & Intel Xeon Gold 6248R, 3.0GHz   & 640GB & NVIDIA A100                     & 40GB     & Ethernet & -     \\ 
\hline
Router                 & Mi Router AX3000       & 1      & Qualcomm IPQ5000 A53, 1.0GHz    & 256MB & -                               & -        & Ethernet & -     \\ 
\hline
\multirow{3}{*}{PC}    & Supermicro X11SCA-F    & 2      & Intel Xeon E-2236, 3.4GHz       & 32GB  & NVIDIA RTX A4000                & 16GB     & Ethernet & 1.0   \\
                       & Supermicro SYS-5038A-I & 2      & Intel Xeon E5-2620 v4, 2.10GHz   & 64GB  & NVIDIA GeForce GTX 1080 Ti      & 12GB * 2 & Wi-Fi    & 1.0   \\
                       & ThinkPad P52s Laptop   & 4      & Intel i5-8350U, 1.70GHz         & 32GB  & NVIDIA Quadro P500              & 2GB      & Wi-Fi    & 0.75  \\ 
\hline
\multirow{3}{*}{Board} & NVIDIA Jetson TX2      & 4      & Dual-Core NVIDIA Denver 2, 2GHz & 8GB   & 256-core NVIDIA Pascal GPU      & 4GB      & Wi-Fi    & 0.75  \\
                       & NVIDIA Jetson Nano     & 4      & ARM Cortex-A57 MPCore, 1.5 GHz  & 4GB   & NVIDIA Maxwell architecture GPU & 2GB      & Wi-Fi    & 0.5   \\
                       & Raspberry Pi 4         & 4      & Quad core Cortex-A72, 1.8GHz    & 8GB   & -                               & -        & Wi-Fi    & 0.25 \\
\bottomrule[1pt]
\end{tabular}
\vspace*{-10pt}
\end{table*}

\vspace{-5pt}
\subsection{Weighted Average Aggregation}
\label{S: M3}

After generating a set of dilated large models, the server aggregates them to obtain the global model. However, we find that directly averaging \cite{DBLP:conf/aistats/McMahanMRHA17} the parameters of all the dilated models leads to severe performance degradation (the accuracy of the aggregated model is only 47.6\% on the MNIST dataset). The reasons are two-folded: 1) the magnitude of the dilated models' parameters varies with their sizes \cite{DBLP:conf/sensys/DengCR0LLZ22}; 2) the parameters of the dilated models through TC operations also carry personalized information from different clients, thus exhibiting distinct patterns and varying skewness toward client-side data distribution (Fig.~\ref{fig:reconstructed_distribution}). Simply aggregating these dilated models overlooks the diverse contributions that heterogeneous clients can make in the aggregation process. 

\textbf{Contribution coordination.} To fuse and balance the diverse personalized information from the dilated models, we first normalize the parameters of all the dilated models to $[0, 1]$ and then assign different learnable \textit{weight vectors} to every network layer in each dilated model for the \textit{weighted aggregation}.
The parameters of the $l$-th aggregated network layer are then expressed as:
\begin{equation}
\small
\setlength{\abovedisplayskip}{0pt}
\setlength{\belowdisplayskip}{0pt}
    \boldsymbol{W}_l = ({\sum_{j=1}^{n}{\boldsymbol{v}_{j,l}\cdot s_j\cdot\boldsymbol{w}_{j,l}}})/\sum_{j=1}^{n}{s_j}
    \label{equ:weighted_average}
\end{equation}
where $\boldsymbol{W}_l$ is the parameters of the $l$-th layer in the aggregated model, $n$ is the number of large models. $\boldsymbol{w}_{j,l}$ and $\boldsymbol{v}_{j,l}$ are the parameters and the corresponding \textit{weight vector} of the $l$-th layer in the $j$-th large model, $s_j$ is the number of data samples that are used for training the $j$-th client model. We iteratively optimize $\boldsymbol{v}_{j,l}$ via gradient descent to gradually balance the distinct contributions.

\textbf{Aggregation enhancement.} To further quantify the different contributions of heterogeneous clients and enhance the aggregation process, we use Kullback-Leibler Divergence (KLD) \cite{lin1991divergence} as a practical criterion to measure the similarity between the parameters of the global model and the dilated models. The higher the similarity, the more contribution the large model will make to the aggregation. Thus, the aggregated global model can attain higher generalizability and a more comprehensive global perspective. The KLD for the $j$-th dilated model is formulated as $KLD_j=\sum_{l=1}^L{\sum_x{\boldsymbol{W}_{G,l}(x)\log{\frac{\boldsymbol{W}_{G,l}(x)}{\boldsymbol{w}_{j,l}(x)}}}}$, where $\boldsymbol{W}_G$ is the global model parameters from the previous communication round. The optimization of the \textit{weight vectors} is expressed as:
\begin{equation}
\small
\setlength{\abovedisplayskip}{0pt}
\setlength{\belowdisplayskip}{0pt}
    \mathcal{L}(\boldsymbol{v}) = \mathcal{L}_\mathcal{D}(\boldsymbol{W}) + \lambda\sum_{j=1}^{n}{KLD_j}
\end{equation}
where $\mathcal{L}$ is the Cross-Entropy loss for the model output, and $\lambda$ is a coefficient for balance. After fine-tuning the \textit{weight vectors}, the aggregated model will be used for the next round.

\vspace{-3pt}
\section{Experiment Setup}

\subsection{Implementation}
We implement \textit{FedConv} with PyTorch \cite{NEURIPS2019_9015} and Flower \cite{DBLP:journals/corr/abs-2007-14390}. The \texttt{load\_state\_dict()} function in PyTorch is overridden to enable the gradient to back-propagate to the \textit{convolution/TC parameters}. We evaluate \textit{FedConv} with a cloud server, a router, and 20 heterogeneous mobile devices with different hardware and network conditions. Detailed configurations of the heterogeneous devices are described in Table \ref{tb:hardware_configuration}. We deploy these edge devices in our offices and laboratories under real-world network conditions.





\vspace{-5pt}
\subsection{Datasets and Models}
\label{S:dataset}
We select two representative mobile applications and use different model architectures and sizes on various datasets.

\textbf{Application\#1: Image Classification.} Imagine classification is a popular computer vision application for FL.
We choose three datasets: 1) \textbf{MNIST} \cite{DBLP:journals/spm/Deng12} consists of 60,000 $28\times28$ gray-scale images of ten handwritten digits. We use a convolutional neural network (CNN) with two convolutional layers and one fully connected layer as the classification model; 2) \textbf{CIFAR10} \cite{krizhevsky2009learning} consists of 60,000 32 $\times$ 32 color images in ten classes. We use ResNet18 \cite{DBLP:conf/cvpr/HeZRS16} to perform the evaluation; 3) \textbf{CINIC10} \cite{darlow2018cinic} contains 180,000 32 $\times$ 32 color images in ten classes. We use GoogLeNet \cite{szegedy2015going} for evaluation.


\textbf{Application\#2: Human Activity Recognition (HAR).} HAR \cite{xu2023practically, 9488802, 9155402, AquaHelper_SenSys23, cui2023mmripple} is realized by analyzing different types of sensor data (\eg, Depth camera, IMU \cite{xu2021limu, liu2021mandipass}, and the channel state information of WIFI signals \cite{ji2023construct, he2023sencom, ji2022sifall}).
We select three datasets: 1) \textbf{WiAR} \cite{DBLP:journals/access/GuoGWLLLFLSY19} contains 480 90 $\times$ 250 Wi-Fi CSI samples of 16 activities. We augment the dataset to 64,000 samples following OneFi \cite{xiao2021onefi}; 2) \textbf{Depth camera dataset} (DCD) \cite{DBLP:conf/mobisys/OuyangXZHX21} contains 5,000 36 $\times$ 36 gray-scale depth images of five common gestures; 3) \textbf{HARBox} \cite{DBLP:conf/mobisys/OuyangXZHX21} captures 9-axis IMU data of five daily activities. A sliding window of 2 seconds is applied to generate 900-dimensional features for each of the 30,000 data samples. The IMU data was collected from 121 users with 77 different smartphones, demonstrating a degree of sensing heterogeneity. As there is no standard model for these datasets, we use a CNN model with three Conv layers and one FC layer that can achieve high accuracy.

\begin{figure}[t]
\setlength{\abovecaptionskip}{-1pt}
    \centering
    \includegraphics[width=0.45\textwidth]{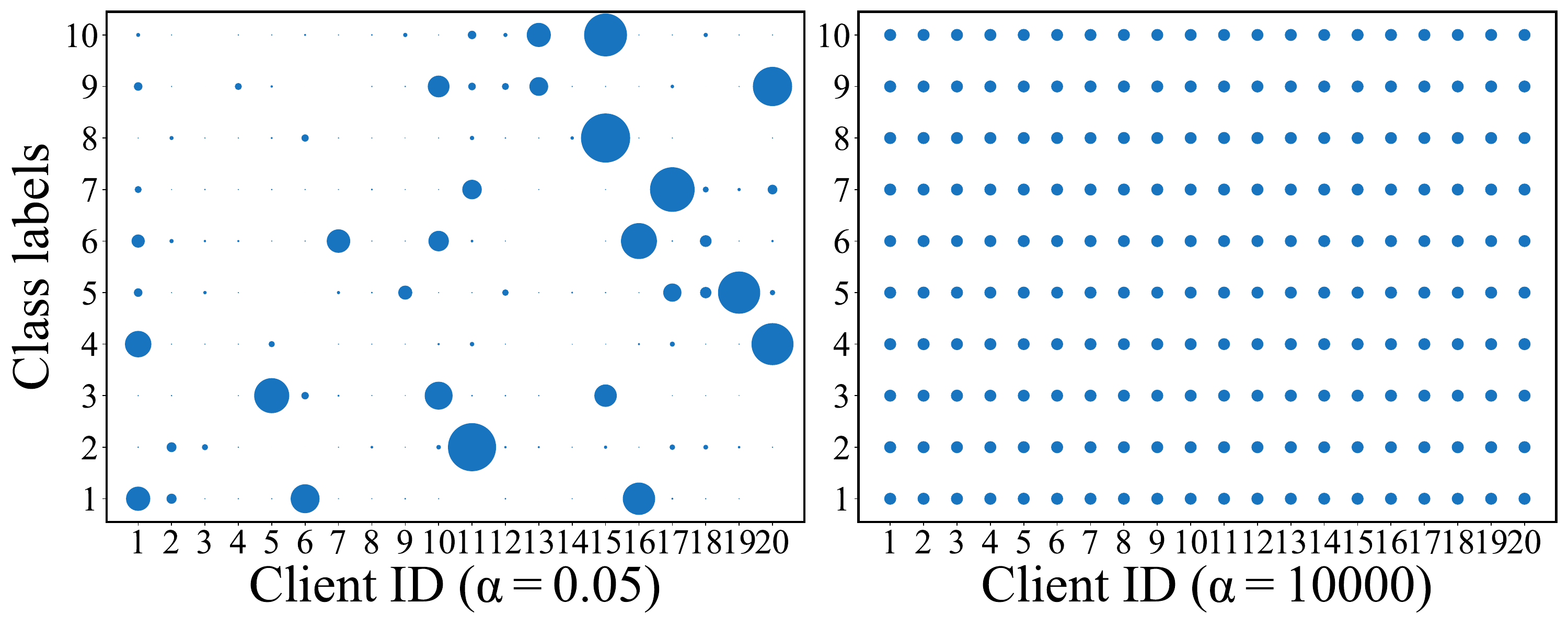}
    \caption{Visualization of Non-IID data. The size of scattered points indicates the number of data samples.}
    \label{fig:class_distribution}
    \vspace{-15pt}
\end{figure}

We divide these datasets into four parts: 1) the IID server-side global data for \textit{convolution}/\textit{TC parameters} and \textit{weight vectors} tuning, 2) IID test data for evaluating the aggregated global model, 3) client-side training and 4) testing data (IID or non-IID). Each part counts for 5\%, 20\%, 70\%, and 5\% of the total dataset, respectively. The first and second parts of the dataset are kept on the server, whereas the third and fourth parts are distributed among heterogeneous clients. Besides, to emulate real-world heterogeneity, we employ different datasets on the server and clients (\S~\ref{sec:case}).


\vspace{-5pt}
\subsection{Baselines}
We compare \textit{FedConv} 
with the following baselines: 1) \textbf{Serveralone} trains one model with only the server-side global data. We evaluate the model using the server-side IID test data and non-IID client-side test data. 2) \textbf{Standalone} allows each client to train an affordable model locally using their private data without parameter exchange. 3) \textbf{FedAvg} \cite{DBLP:conf/aistats/McMahanMRHA17} is a classic FL paradigm where clients collaboratively train a shared global model and upload the updated model parameters to a central server for averaging aggregation. Due to the constrained resources of some devices, we assign the smallest affordable models to all clients. 4) \textbf{FedMD} \cite{DBLP:journals/corr/abs-1910-03581} utilizes knowledge distillation to reach a consensus among heterogeneous client models through training on a public dataset. 5) \textbf{LotteryFL} \cite{DBLP:conf/ieeesec/LiSWDLCL21} generates sub-models by exploiting the Lottery Ticket hypothesis on heterogeneous clients for personalization. 6) \textbf{Hermes} \cite{DBLP:conf/mobicom/0005SLPLC21} finds a sparse sub-model for each client by using a channel-wise pruning scheme to reduce the communication overhead. 7) \textbf{TailorFL} \cite{DBLP:conf/sensys/DengCR0LLZ22} produces sub-models by filter-level pruning based on the learned importance value of each filter. 8) \textbf{HeteroFL} \cite{DBLP:conf/iclr/Diao0T21} is a parameter sharing method that allows each client to select a subset of the parameters from the global model. 9) \textbf{FedRolex} \cite{alam2022fedrolex} adopts dynamic rolling windows when extracting sub-models for heterogeneous clients.


\begin{figure*}[t]
\vspace{-8pt}
\setlength{\abovecaptionskip}{-5pt}
\subfigtopskip=-5pt
\subfigcapskip=-5pt
    \centering
    \subfigure[Global model accuracy comparison]{
        \includegraphics[width=0.45\textwidth]{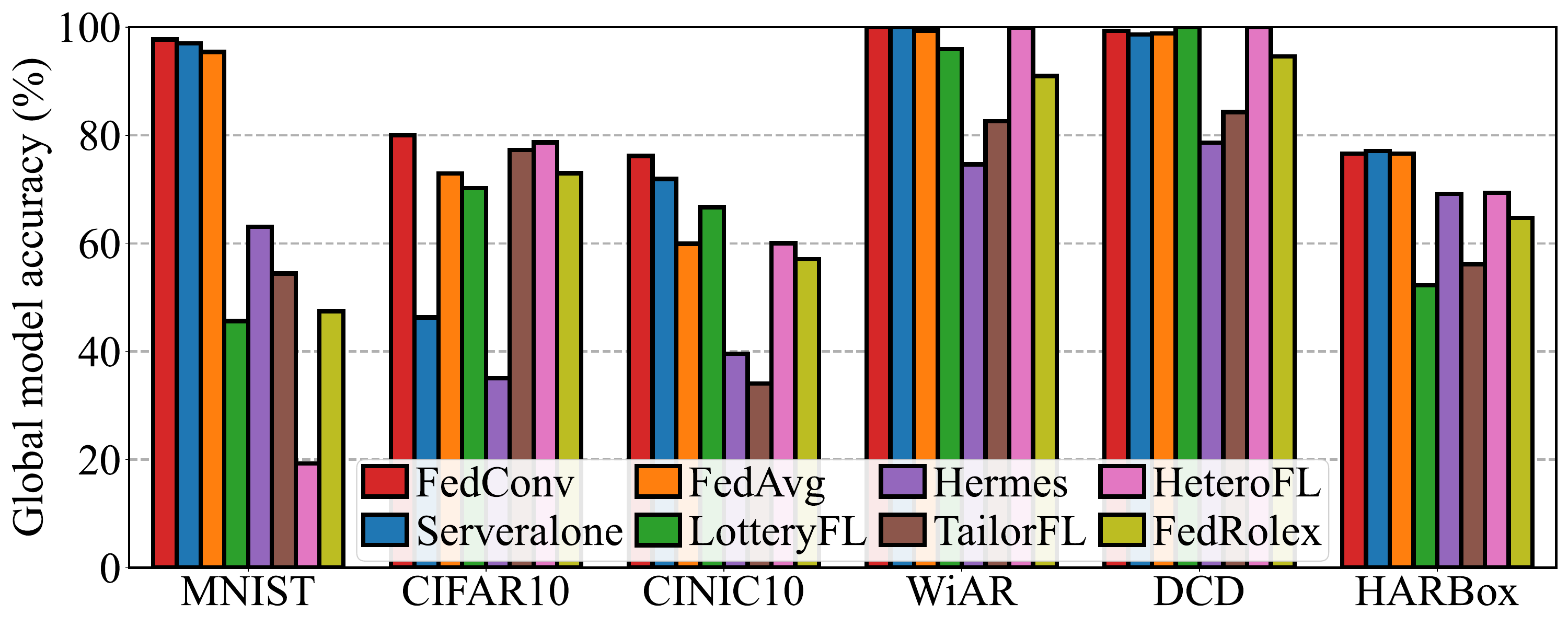}
        \label{fig:overall_accuracy_global}
    }
    \centering
    \subfigure[Client model accuracy comparison]{
        \includegraphics[width=0.45\textwidth]{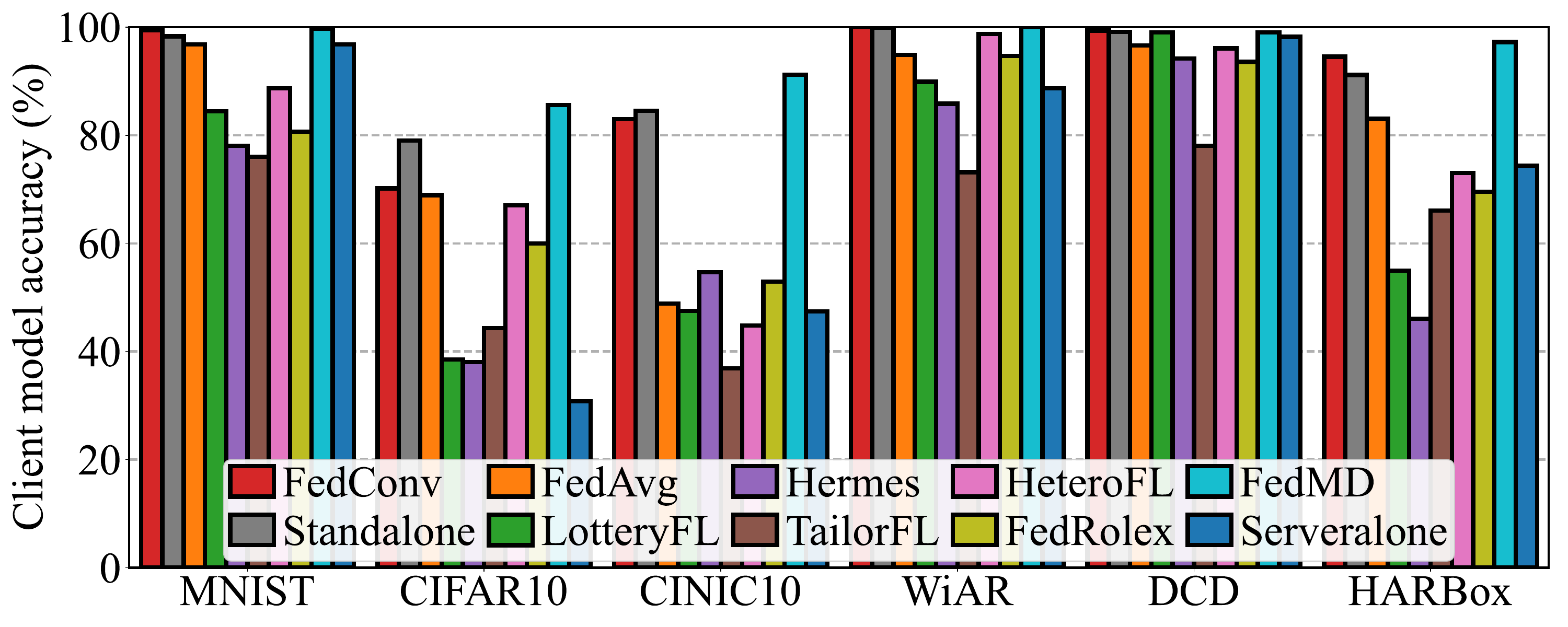}
        \label{fig:overall_accuracy_client}
    }
    \caption{Accuracy comparison under model heterogeneity ($\alpha=0.1$).}
    \vspace{-8pt}
\end{figure*}

\begin{figure*}[t]
\setlength{\abovecaptionskip}{-5pt}
\subfigtopskip=-5pt
\subfigcapskip=-7pt
    \centering
    \includegraphics[width=0.75\textwidth]{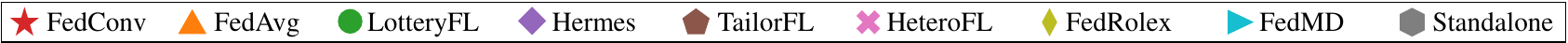}\vspace{8pt}
    \centering
    \subfigure[Global model accuracy comparison]{
        \includegraphics[width=0.99\textwidth]{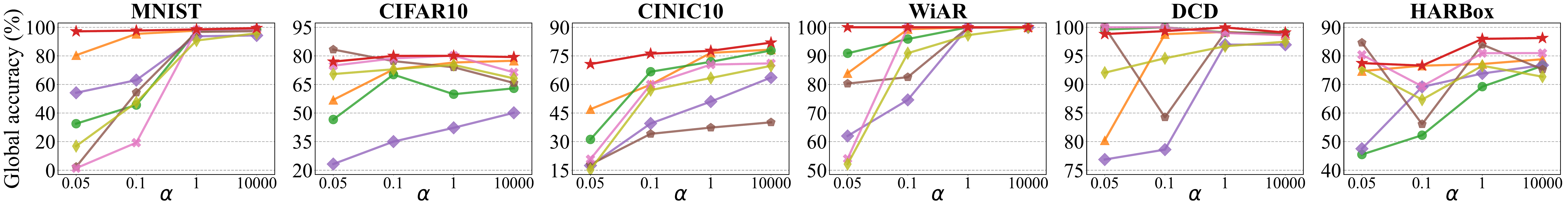}
        \label{fig:global_accuracy_all}
    }
    \centering
    \subfigure[Client model accuracy comparison]{
        \includegraphics[width=0.99\textwidth]{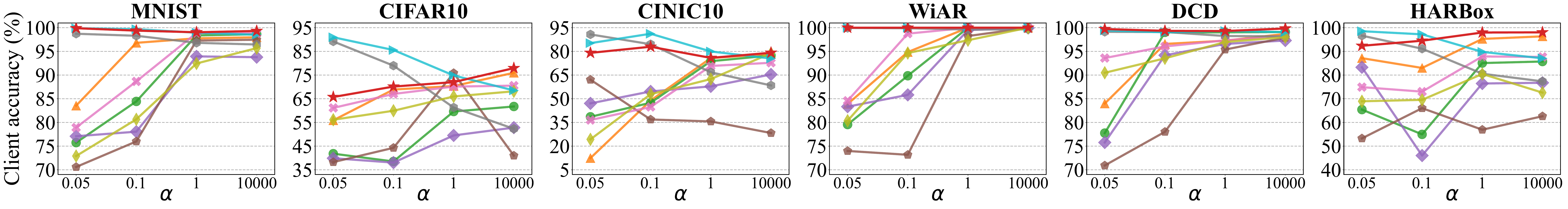}
        \label{fig:client_accuracy_all}
    }
    \caption{The inference accuracy of aggregated global models and client models on different datasets.}
    \label{fig:inference_accuracy}
    \vspace{-12pt}
\end{figure*}

\vspace{-5pt}
\subsection{Heterogeneity Consideration}
For model heterogeneity, we consider four SRs: 0.25, 0.5, 0.75, and 1.0, according to the resource profiles of the heterogeneous clients. The SR for each client is detailed in Table \ref{tb:hardware_configuration}. For powerful clients, we assign larger SRs (\eg, 0.75 for laptops), and for resource-constrained clients, we assign smaller SRs (\eg, 0.25 for Raspberry Pis). For data heterogeneity, we sample the disjoint non-IID client-side data using the Dirichlet distribution $\mathbf{Dir}(\alpha)$. A larger $\alpha$ (\eg, $10000$) indicates a more homogeneous distribution and a smaller $\alpha$ (\eg, $0.05$) generates a more heterogeneous distribution \cite{DBLP:journals/corr/abs-1909-06335}, The sample distribution among different classes is illustrated in Fig. \ref{fig:class_distribution}. 

\vspace{-5pt}
\subsection{Hyper-parameter Settings}
For baselines and \textit{FedConv}, we set the number of communication rounds to 100. Each client performs 5 local training epochs with a learning rate of 0.001. In the model compression and dilation process, the stride and padding of all the convolution parameters are 1 and 0. The server-side pre-training epoch number is 5. The epoch number for updating \textit{convolution/TC parameters} are both 20, and the $T_{max}$, $lr_{min}$, $lr_{max}$ in the cosine annealing scheduler are 4, 0.00001, and 0.001, respectively. $s_p$ and $s_n$ (Fig. \ref{fig:modified_activation}) in the activation function are 0.85 and 0.001, respectively. In model aggregation, the number of epochs, the learning rate for updating weight vectors, and $\lambda$ in Eq. (\ref{equ:weighted_average}) are 10, 0.001, and 0.2, respectively.

\vspace{-5pt}
\section{Evaluation}

\subsection{Metrics}

\textbf{Training performance}: 1) \textit{Inference accuracy}: we measure the global model accuracy with the server-side test dataset to evaluate the generalizability of the global model. We also report the average client model accuracy with client-side private test datasets to evaluate the effectiveness of personalization. 2) \textit{Communication cost}: we use the \textit{Pympler} library to monitor the network traffic of all the clients over 100 communication rounds.

\noindent\textbf{Runtime performance}: 1) \textit{Memory footprint}: the real-time GPU memory usage is monitored using the PyTorch CUDA Toolkit. We track each client's process ID over 100 communication rounds to monitor their CPU usage and report the average value. 2) \textit{Wall-clock time}: we measure the execution time of each client from receiving model parameters to finishing the training task, and report the average wall-clock time in each round. 

\vspace{-5pt}
\subsection{Overall Performance}

We evaluate the overall performance of \textit{FedConv} with heterogeneous models and data distribution.

\vspace{-5pt}
\subsubsection{Global model performance.}
We first evaluate the accuracy of the aggregated global model to demonstrate its generalizability. Standalone and FedMD are excluded because they do not create global models. Fig. \ref{fig:overall_accuracy_global} shows the global model accuracy under the same degree of heterogeneous data ($\alpha=0.1$). Serveralone achieves a higher global model accuracy than the baselines in most cases, as the server-side data for training and testing are both IID. \textit{FedConv} achieves average improvements of 20.5\%, 13.8\%, and 10.5\% compared with pruning-based methods (Hermes and TailorFL), parameter sharing-based method (HeteroFL and FedRolex) and other baselines (FedAvg and LotteryFL), respectively. Since we assign the smallest affordable model to all clients in FedAvg, the client models have an insufficient number of parameters for training. Therefore, \textit{FedConv} can outperform FedAvg even with IID data. This shows the superior generalization performance of \textit{FedConv}.

\begin{table*}[htbp]
\vspace{-10pt}
\centering
\setlength{\abovecaptionskip}{-1pt}
    \footnotesize
    \caption{System resource overhead.}
    \label{tb:system_resource}
    \begin{tabular}{cccccccc@{\hspace{0.5cm}}cccccc} 
    \toprule[1pt]
\multirow{2}{*}{\textbf{Metric} }                                                        & \multirow{2}{*}{\textbf{System} } & \multicolumn{6}{c}{\textbf{Heterogeneous Data ($\alpha=0.05$)}}                                                 & \multicolumn{6}{c}{\textbf{Homogeneous Data ($\alpha=10000$)}}                                                \\ 
\cline{3-14}
                                                                                &                          & \textbf{MNIST}          & \textbf{CIFAR10}          & \textbf{CINIC10}          & \textbf{WiAR}           & \textbf{DCD}            & \textbf{HARBox}        & \textbf{MNIST}         & \textbf{CIFAR10}        & \textbf{CINIC10}          & \textbf{WiAR}           & \textbf{DCD}            & \textbf{HARBox}         \\ 
\toprule[1pt]
\multirow{8}{*}{\begin{tabular}[c]{@{}c@{}}\textbf{Memory}\\\textbf{Footprint}\\\textbf{CPU + GPU}\\\textbf{(GB)}\end{tabular}} & Standalone               & 2.14           & 3.51             & 4.07             & 3.95           & 2.24           & 2.19          & 2.13          & 3.47           & 4.47             & 4.03           & 2.21           & 2.17           \\
                                                                                & FedAvg                   & 1.90           & 2.40             & 3.31             & 2.39           & 1.98           & 2.01          & 1.90          & 2.51           & 2.79             & 2.36           & 1.88           & 2.08           \\
                                                                                & FedMD                    & 2.71           & 3.65             & 7.51             & 4.71           & 2.99           & 2.79          & 2.71          & 3.65           & 7.93             & 4.58           & 2.99           & 2.81           \\
                                                                                & LotteryFL                & 2.62           & 3.51             & 4.30             & 3.23           & 2.69           & 2.67          & 2.63          & 3.49           & 4.36             & 3.27           & 2.70           & 2.66           \\
                                                                                & Hermes                   & 2.64           & 3.45             & 6.07             & 3.28           & 2.73           & 2.69          & 2.64          & 3.35           & 6.13             & 3.32           & 2.72           & 2.68           \\
                                                                                & TailorFL                 & 2.75           & 3.61             & 5.09             & 3.41           & 2.79           & 2.71          & 2.75          & 3.47           & 7.52             & 3.16           & 2.77           & 2.70           \\
                                                                                & HeteroFL                 & 2.63           & 3.31             & 4.15             & 3.25           & 2.73           & 2.67          & 2.63          & 3.45           & 4.10             & 3.08           & 2.73           & 2.67           \\
                                                                                &FedRolex& 2.63  & 3.21    & 4.15    & 3.25  & 2.72  & 2.67 & 2.60 & 3.54  & 4.16    & 3.16  & 2.68  & 2.69  \\
                                                                                &\textbf{\textit{FedConv}}& \textbf{2.52}  & \textbf{3.21}    & \textbf{4.15}    & \textbf{3.02}  & \textbf{2.60}  & \textbf{2.67} & \textbf{2.52} & \textbf{3.35}  & \textbf{4.10}    & \textbf{3.14}  & \textbf{2.62}  & \textbf{2.67}  \\ 
\hline
\multirow{8}{*}{\begin{tabular}[c]{@{}c@{}}\textbf{Wall-clock}\\\textbf{ Time (s)}\end{tabular}}   & Standalone               & 3.87           & 24.65            & 279.62           & 8.05           & 5.91           & 3.54          & 9.38          & 52.38          & 273.52           & 7.60           & 6.14           & 3.56           \\
                                                                                & FedAvg                   & 7.05           & 39.19            & 285.30           & 10.62          & 10.19          & 10.09         & 13.75         & 97.95          & 1711.34          & 20.79          & 43.67          & 26.98          \\
                                                                                & FedMD                    & 44.34          & 437.14           & 5370.83          & 55.03          & 75.25          & 32.92         & 45.17         & 475.42         & 6700.17          & 64.43          & 79.10          & 34.53          \\
                                                                                & LotteryFL                & 9.18           & 147.98           & 699.35           & 8.89           & 8.61           & 5.69          & 17.59         & 235.89         & 1829.33          & 19.77          & 22.06          & 10.92          \\
                                                                                & Hermes                   & 43.22          & 714.00           & 5580.71          & 103.90         & 169.97         & 104.53        & 43.84         & 937.82         & 7621.38          & 117.85         & 217.97         & 115.31         \\
                                                                                & TailorFL                 & 6.98           & 62.89            & 393.46           & 14.44          & 12.72          & 10.11         & 13.61         & 99.60          & 813.94           & 25.53          & 13.96          & 13.27          \\
                                                                                & HeteroFL                 & 6.96           & 42.56            & 641.21           & 10.78          & 10.03          & 5.10          & 13.56         & 82.07          & 1310.81          & 22.26          & 23.90          & 10.98          \\
                                                                                & FedRolex                 & 6.92           & 45.98            & 602.48           & 11.57          & 12.34         & 4.87          & 12.46       & 84.25          & 1389.41          & 23.64          & 20.14         & 11.26          \\
                                                                                & \textbf{\textit{FedConv}}         & \textbf{5.96}  & \textbf{40.68}   & \textbf{264.30}  & \textbf{12.96} & \textbf{10.15} & \textbf{4.40} & \textbf{10.33} & \textbf{71.26} & \textbf{1406.87} & \textbf{21.79} & \textbf{17.22} & \textbf{9.89}  \\ 
\bottomrule[1pt]
\end{tabular}
\vspace{-10pt}
\end{table*}

\begin{table}
\vspace{-2pt}
\centering
\footnotesize
\setlength{\abovecaptionskip}{-1pt}
\caption{Communication overhead comparison (GB).}
\label{tb:communication}
\begin{tabular}{ccccccc}
\toprule[1pt]
\textbf{System} & \textbf{MNIST} & \textbf{CIFAR10} & \textbf{CINIC10} & \textbf{WiAR} & \textbf{DCD} & \textbf{HARBox}\\
\toprule[1pt]
FedAvg                  & 14.80 & 4815.84 & 2697.85 & 28.24 & 13.45 & 8.87    \\
FedMD                   & 19.99 & 5126.46 & 2859.79 & 40.91 & 19.94 & 16.24   \\
LotteryFL               & 11.11 & 4713.91 & 2623.93 & 23.01 & 10.05 & 8.55    \\
Hermes                  & 16.34 & 7099.66 & 2848.83 & 36.63 & 15.02 & 12.95   \\
TailorFL                & 11.40 & 4787.18 & 2686.15 & 24.30 & 10.32 & 8.82    \\
HeteroFL                & 11.11 & 4713.91 & 2623.93 & 23.01 & 10.05 & 8.55    \\
FedRolex       & 11.11 & 4713.91 & 2623.93 & 23.01 & 10.05 & 8.55
\\\textbf{\textit{FedConv}}                  & \textbf{11.11}  & \textbf{4713.91} & \textbf{2623.93} & \textbf{23.01} & \textbf{10.05} & \textbf{8.55}    \\
\bottomrule[1pt]
\end{tabular}
\vspace{-15pt}
\end{table}

Moreover, Fig. \ref{fig:global_accuracy_all} shows the global model accuracy of \textit{FedConv} and all baselines across different data heterogeneity on all datasets. We can see that the performance enhancement of \textit{FedConv} becomes more significant as $\alpha$ decreases, meaning that \textit{FedConv} can better cope with the increased data heterogeneity. Although \textit{FedConv} does not obviously outperform FedAvg with homogeneous data, it exhibits better generalizability and robustness in the global model under heterogeneous data. \textit{FedConv} also provides better personalization performance for clients (\S~\ref{S:client_model_performance}). The performance improvements of the global model stem from our \textit{convolutional compression} and \textit{TC dilation} methods. They facilitate the information embedded in the global model being preserved and transferred from the server to clients through our \textit{learning-on-model} approach.



\vspace{-5pt}
\subsubsection{Client model performance.}
\label{S:client_model_performance}
To evaluate the personalization performance, we measure the accuracy of each client model with client-side test datasets and report the average value. Fig. \ref{fig:overall_accuracy_client} shows that with the same heterogeneous data settings, \textit{FedConv} outperforms baselines (FedAvg, LotteryFL, Hermes, TailorFL, HeteroFL, and FedRolex) with accuracy improvements ranging from 8.4\% to 50.6\%. In Serveralone, when evaluating the global model using the client-side non-IID data, the accuracy of the client model drops below that of most baseline systems. This is because, in \textit{FedConv}, the server-side data occupies a small portion (5\%) of the entire dataset. Therefore, Serveralone's global model hasn't seen sufficient data, leading to degraded performance on the client-side non-IID data.


Additionally, Fig. \ref{fig:client_accuracy_all} shows the client model accuracy of \textit{FedConv} and all baselines with different data heterogeneity. We can see that the performance disparities become more substantial as $\alpha$ decreases, implying that \textit{FedConv} is more robust and can achieve consistently high accuracy across diverse data distribution. This performance gain stems from the TC dilation process, where distinct \textit{TC parameters} are assigned to each uploaded client model on the server. The rescaled large models will thereby preserve the personalization information from clients, which is then aggregated into the global model. Besides, Fig.~\ref{fig:overall_accuracy_client} shows that, with sensing heterogeneity in the HARBox dataset, \textit{FedConv} achieves a better and more stable performance. However, when $\alpha$ is small (\eg, $\alpha\in\{0.05, 0.1\}$ on CIFAR10), the client model accuracy of FedMD is higher than \textit{FedConv}. The better performance stems from the distilled knowledge shared by all clients. Nonetheless, the downside is that it imposes excessive communication and computational overhead on clients (Table~\ref{tb:system_resource} \& Table~\ref{tb:communication}). By contrast, \textit{FedConv} can achieve comparable personalization performance without an extra burden on clients. In practice, we can further improve the personalization performance by adding task-specific layers \cite{DBLP:journals/corr/abs-2207-08147} (detailed in\S~\ref{S: ablation}). 


\textbf{Remarks.} \textit{FedConv} exhibits significant performance gains in both global and client models across various settings. The parameter information of the global model can be preserved via our \textit{convolutional compression} module. We suspect that the performance instability of some baselines might be attributed to the information loss in model pruning and the imbalance issue in parameter sharing. 

\begin{figure*}[t]
\vspace{-10pt}
    \centering
    \begin{minipage}[t]{0.49\textwidth}
        \setlength{\abovecaptionskip}{-4pt}
        \subfigtopskip=-4pt
        \subfigcapskip=-4pt
        \centering
        \subfigure[\textit{FedConv} on different datasets]{
            \includegraphics[width=0.45\textwidth]{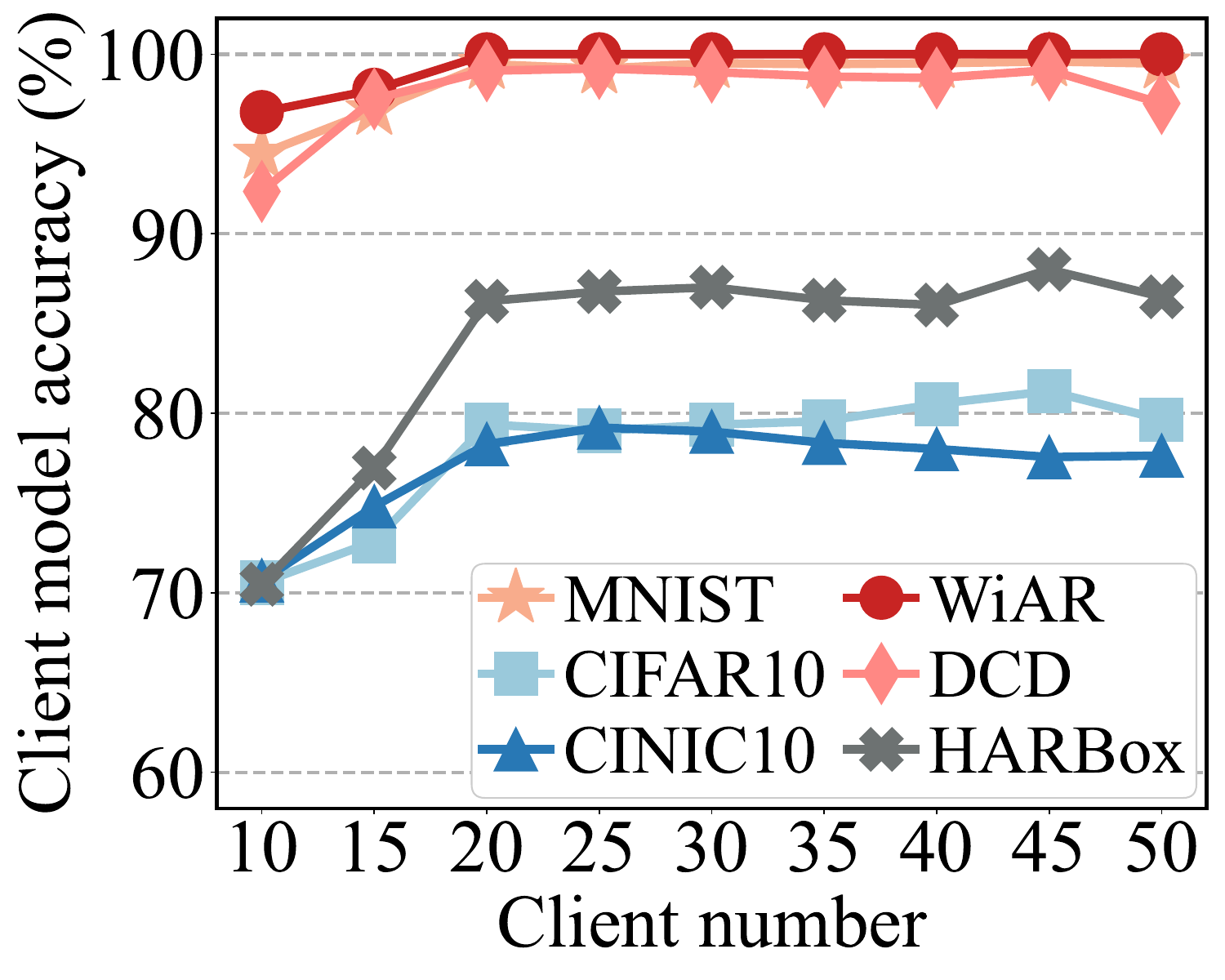}
            \label{fig:varying_client_number_FedConv}
        }
        \centering
        \subfigure[\textit{FedConv} vs. baselines (CIFAR10)]{
            \includegraphics[width=0.45\textwidth]{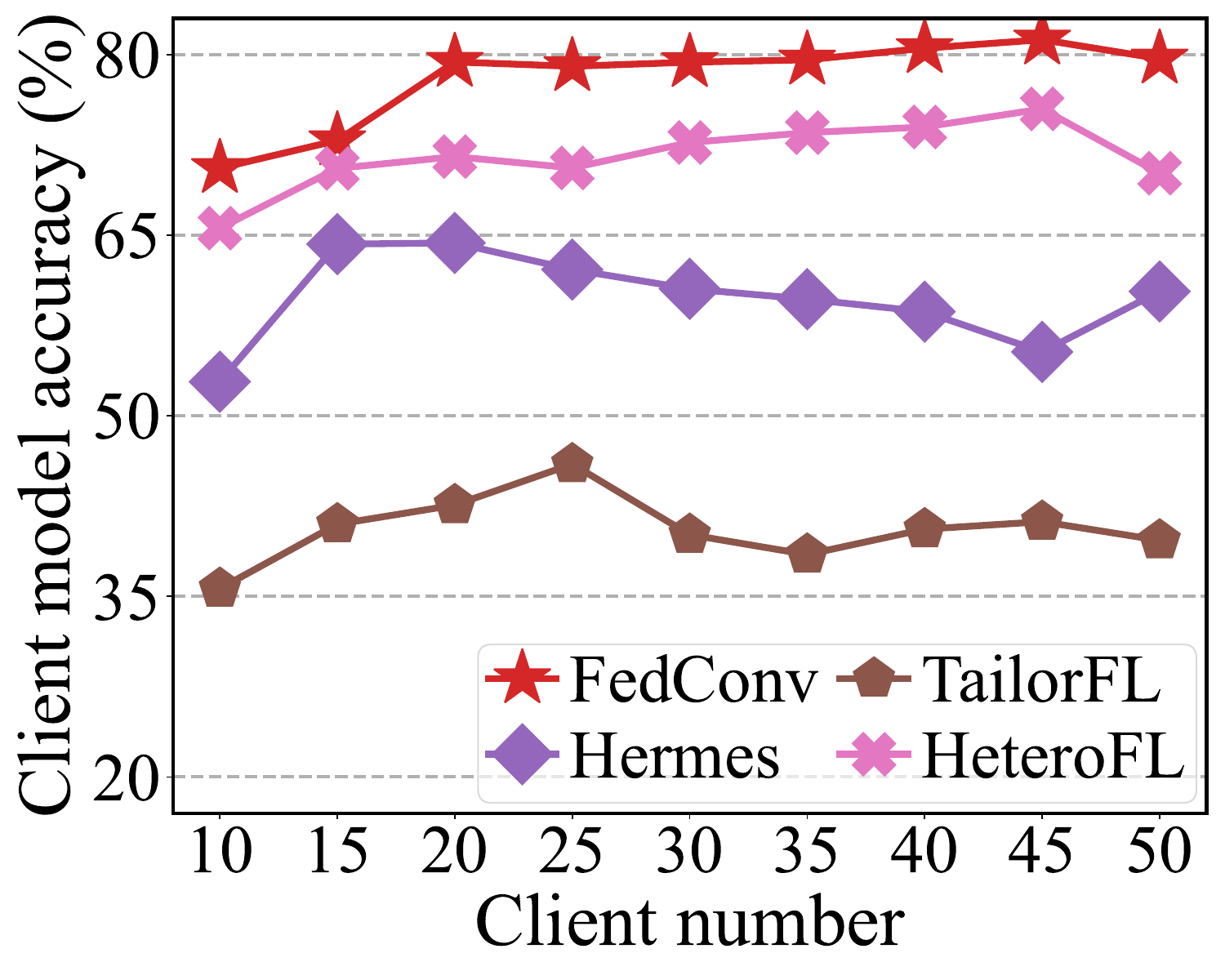}
            \label{fig:varying_client_number_CIFAR}
        }
        \caption{Varying number of clients.}
    \end{minipage}
    \begin{minipage}[t]{0.49\textwidth}
        \setlength{\abovecaptionskip}{-4pt}
        \subfigtopskip=-4pt
        \subfigcapskip=-4pt
        \centering
        \subfigure[\textit{FedConv} on different datasets]{
            \includegraphics[width=0.45\textwidth]{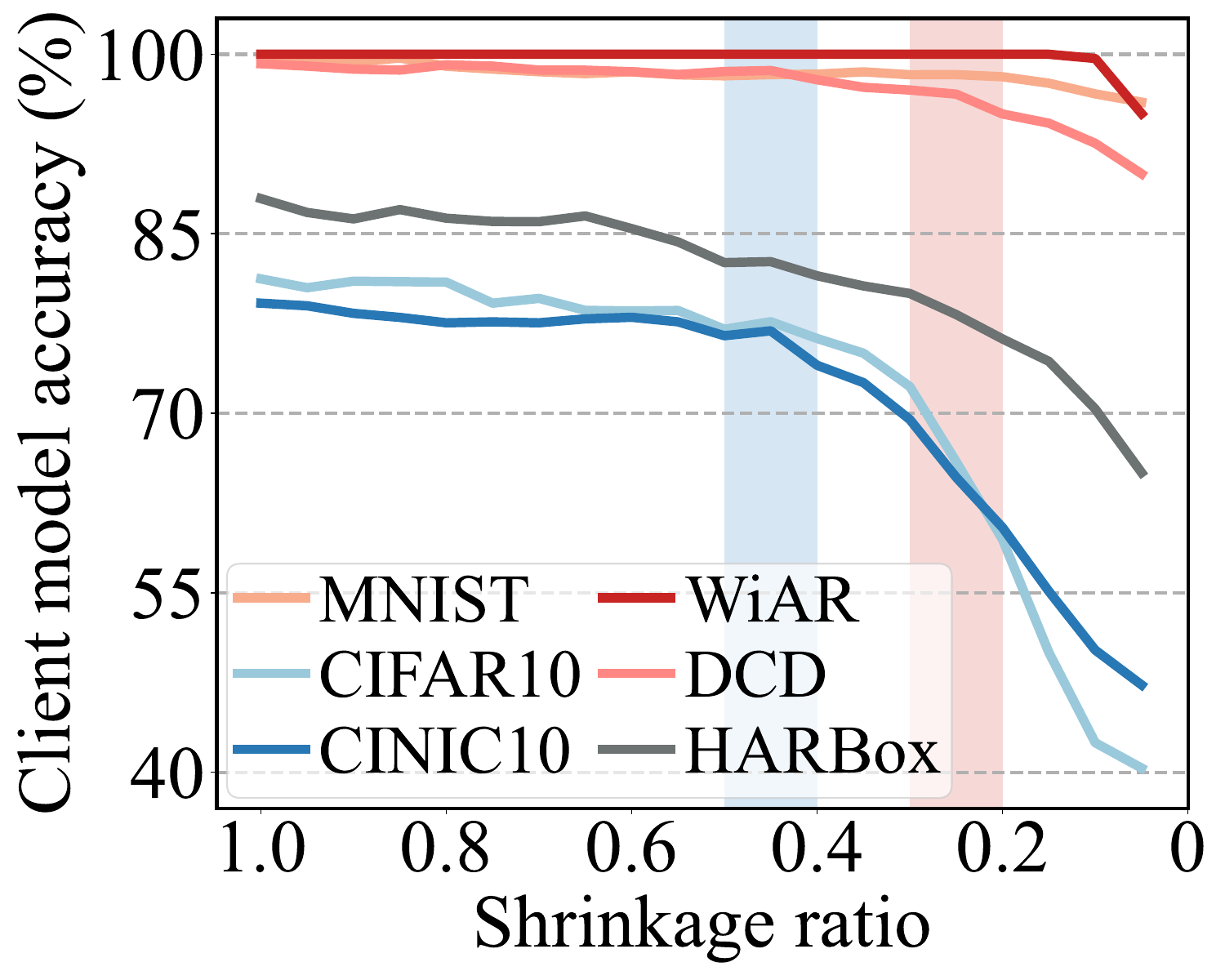}
            \label{fig:varying_SR_FedConv}
        }
        \centering
        \subfigure[\textit{FedConv} vs. baselines (CIFAR10)]{
            \includegraphics[width=0.45\textwidth]{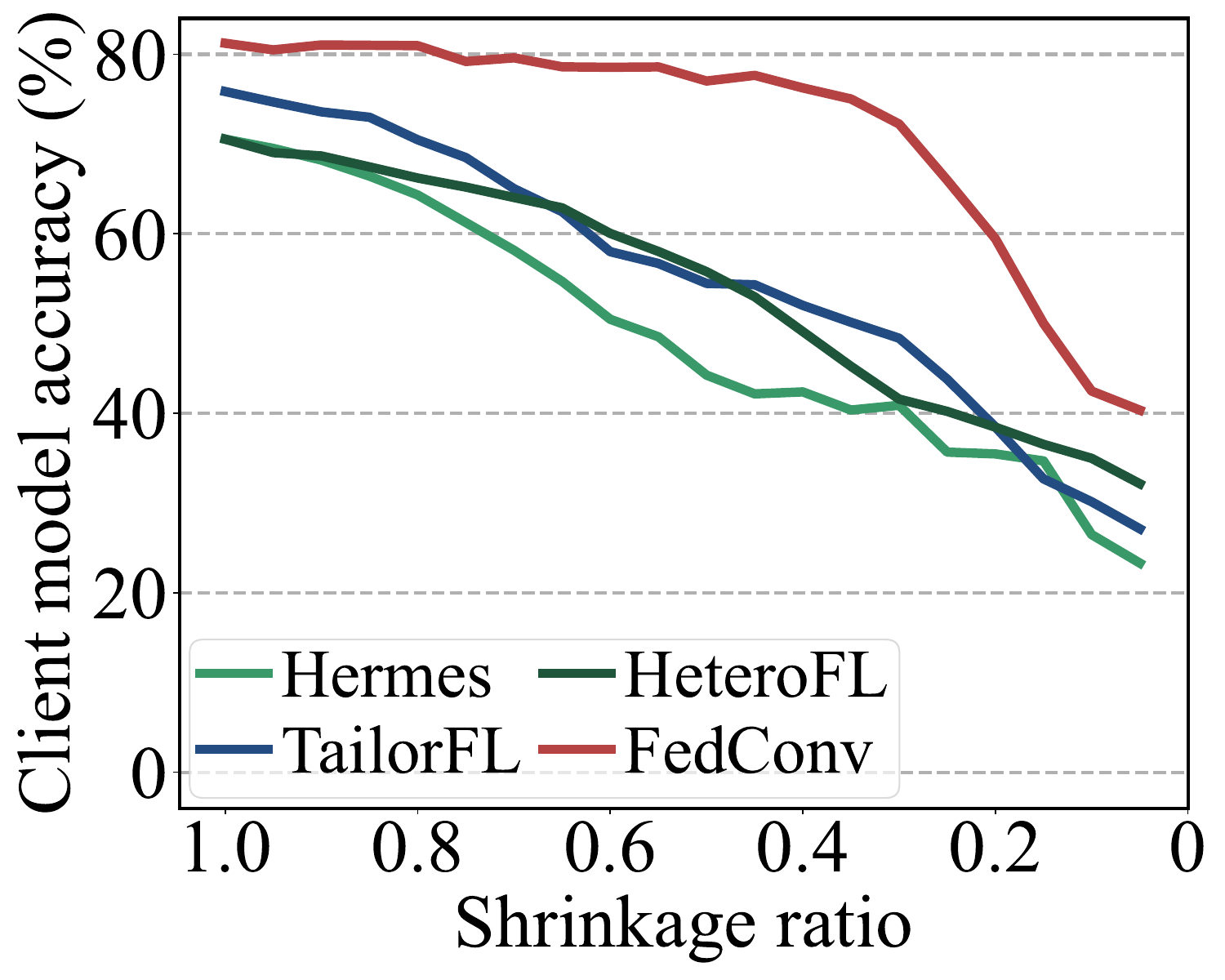}
            \label{fig:varying_SR_CIFAR}
        }
        \caption{Varying shrinkage ratios.}
    \end{minipage}
    \vspace{-5pt}
\end{figure*}

\begin{figure*}[t]
\setlength{\abovecaptionskip}{-5pt}
\subfigtopskip=-4pt
\subfigcapskip=-4pt
    \centering
    \subfigure[Varying server data size]{
        \includegraphics[width=0.225\textwidth]{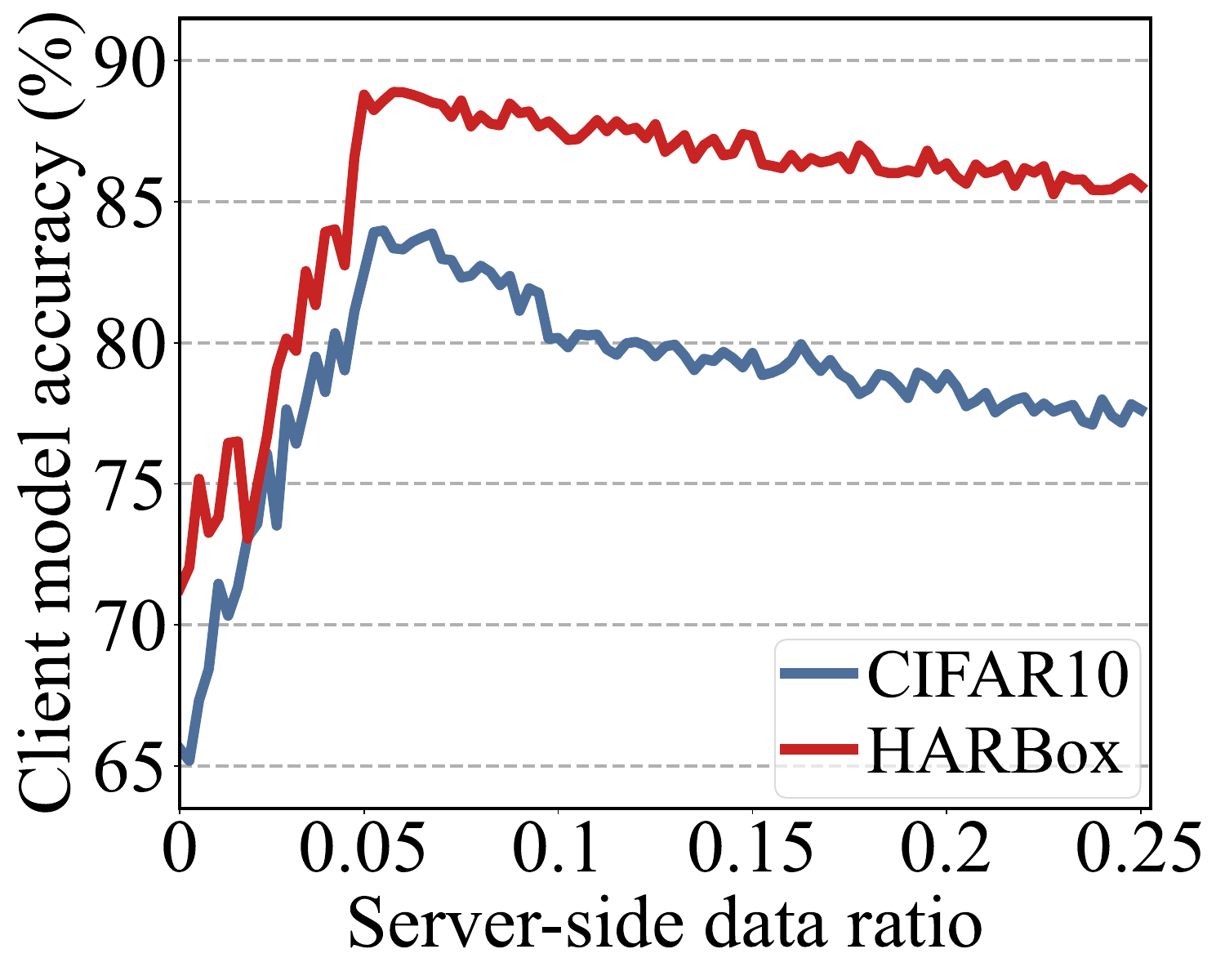}
        \label{fig:varying_server_data}
    }
    \centering
    \subfigure[Varying convolution/TC epochs]{
        \includegraphics[width=0.225\textwidth]{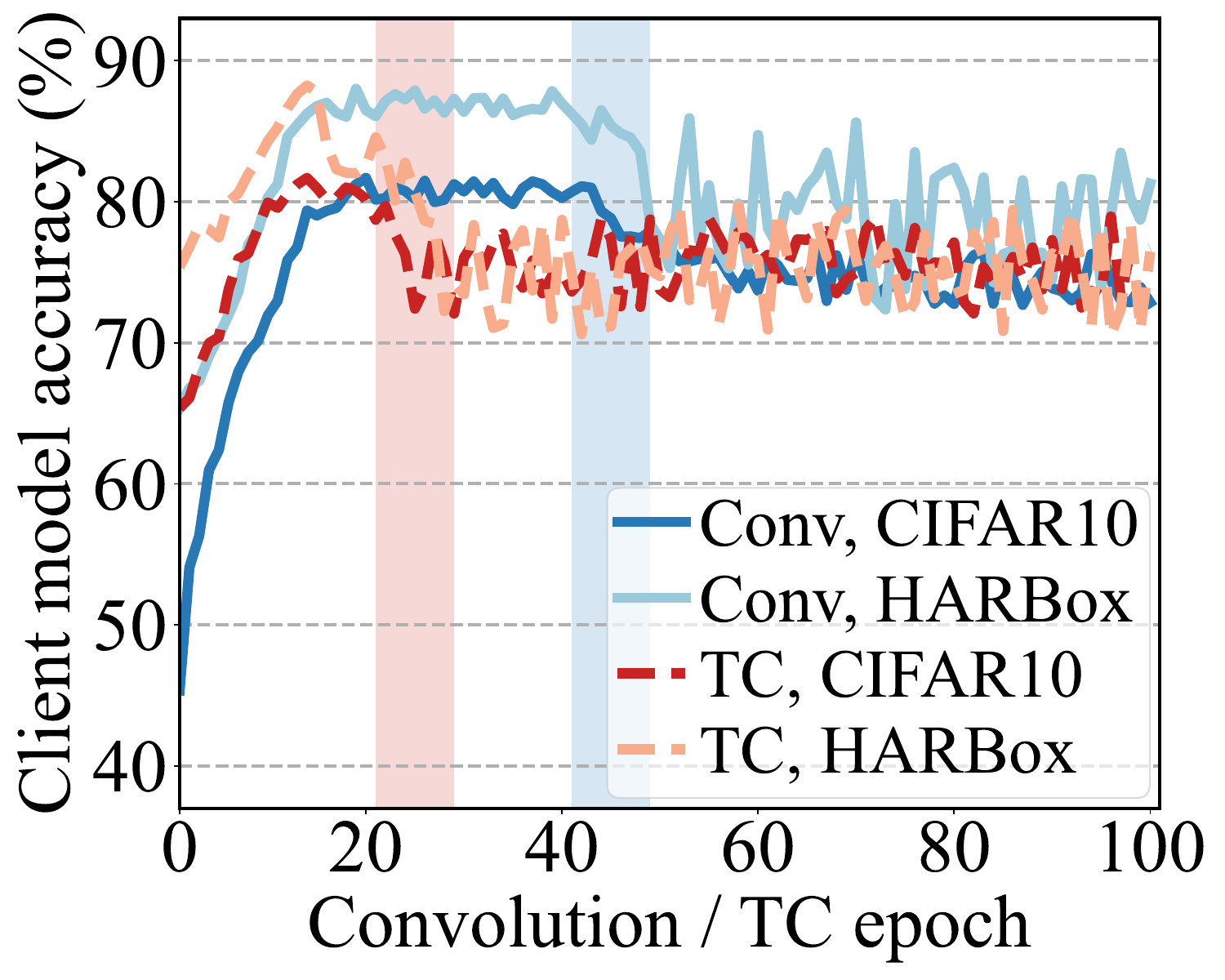}
        \label{fig:varying_conv_epoch}
    }
    \centering
    \subfigure[Varying aggregation epochs]{
        \includegraphics[width=0.225\textwidth]{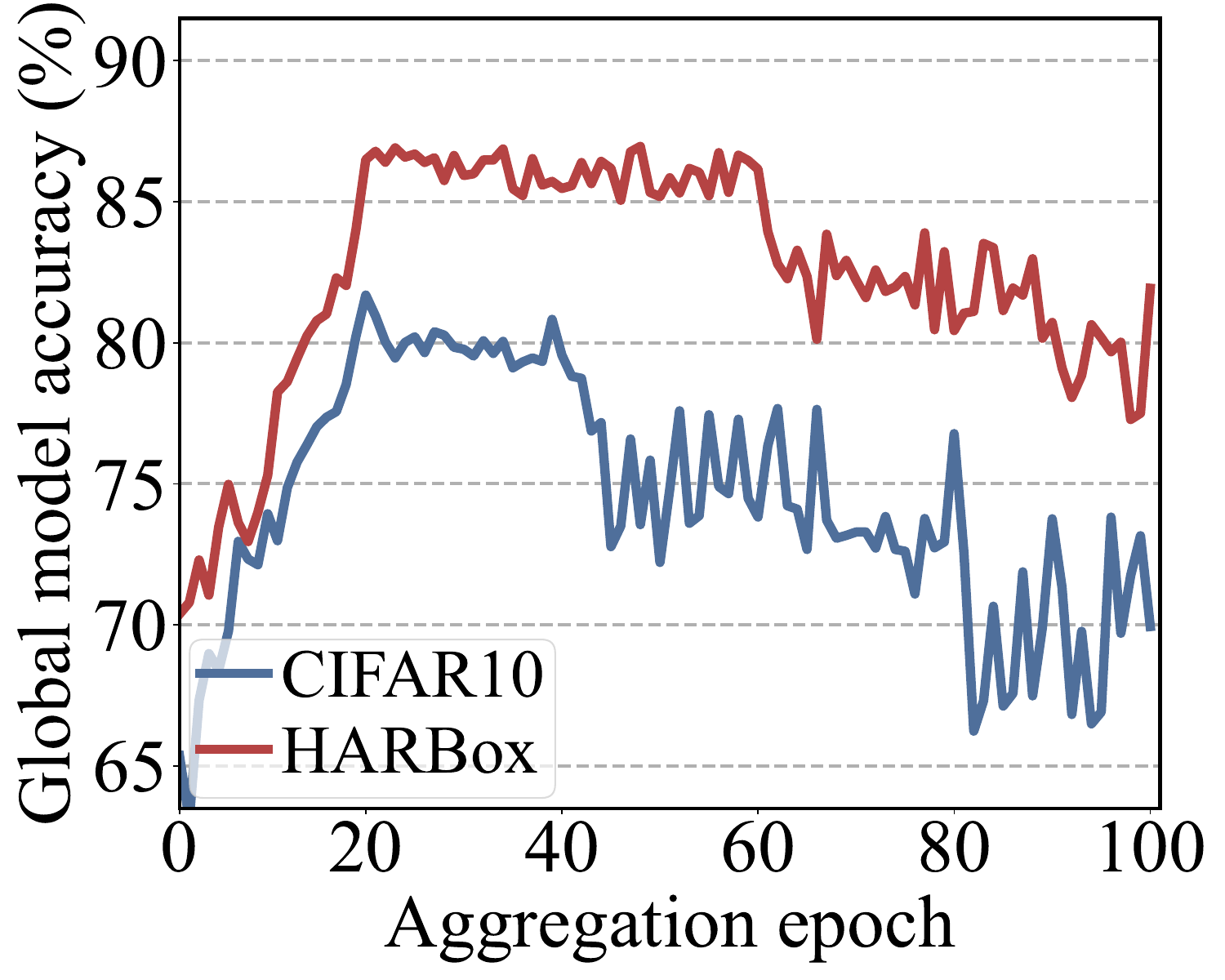}
        \label{fig:varying_aggregation}
    }
    \centering
    \subfigure[Client model accuracy (CIFAR10)]{
        \includegraphics[width=0.225\textwidth]{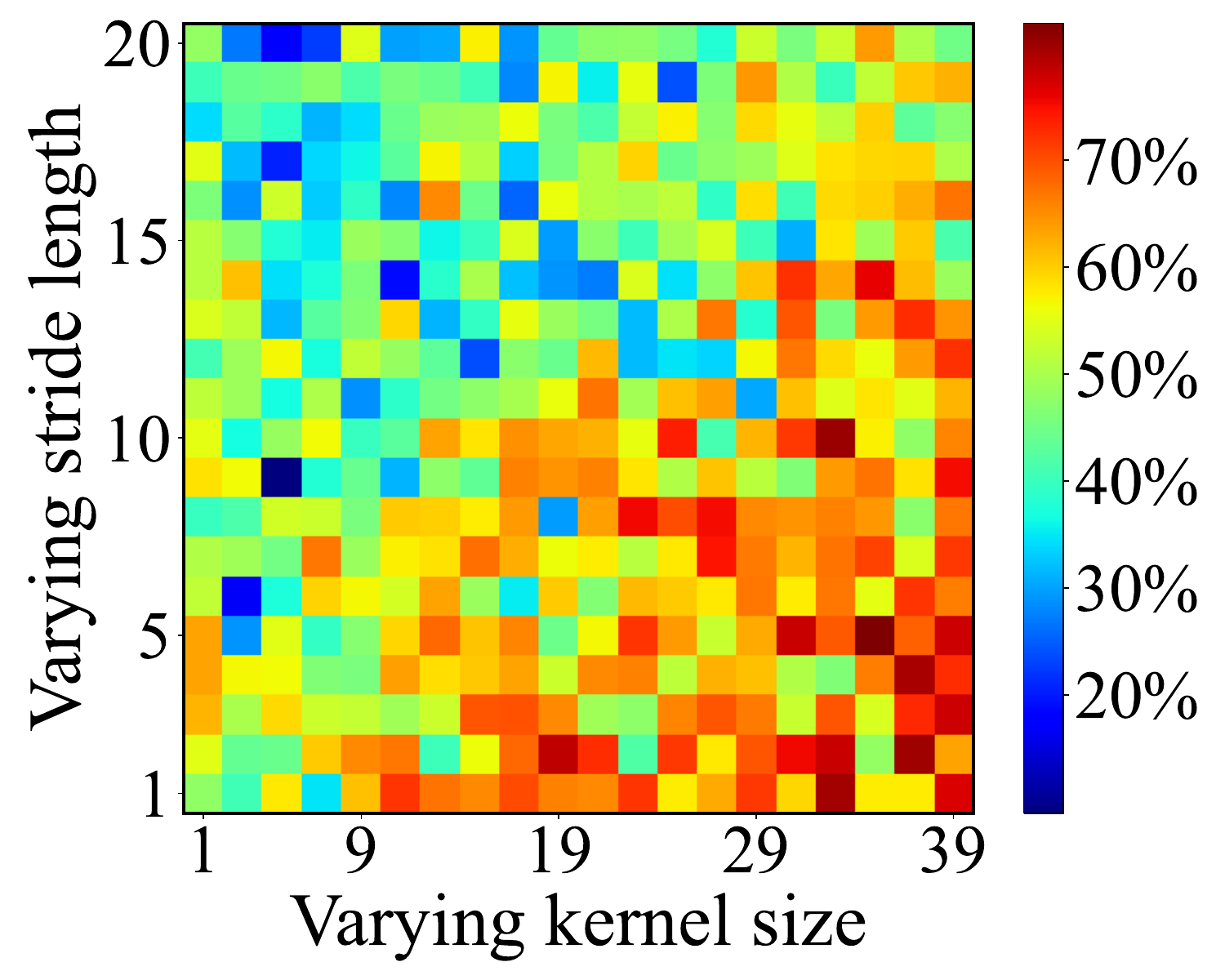}
        \label{fig:varying_conv_pattern_CIFAR10}
    }
    \caption{Sensitivity analysis of varying hyper-parameters in \textit{FedConv}.}
    \label{fig:sensitivity_analysis}
    \vspace{-15pt}
\end{figure*}

\vspace{-5pt}
\subsection{Overhead Assessment}
\label{S:overhead}
We evaluate the memory footprint, wall-clock time, and communication overhead of each client in \textit{FedConv} and baselines with both homogeneous ($\alpha$=10000) and heterogeneous ($\alpha$=0.05) data across clients. Table \ref{tb:system_resource} provides an overview of the average memory usage and the average wall-clock time of each client. With the same set of SRs, \textit{FedConv} achieves an average saving of 40.6\% in memory cost and 54.6\% in computation overhead compared with the baselines, respectively. Furthermore, when the client model is complex (ResNet18 and GoogLeNet), \textit{FedConv} only needs approximately half of the memory and training time compared to the pruning-based methods. For example, in the homogeneous data condition, \textit{FedConv} needs 2GB less memory and saves around 90 minutes of wall-clock time than Hermes in one single round. This is because the computation-intensive pruning operations are executed on the resource-constrained clients. In contrast, clients in \textit{FedConv} only need to perform local training in each round, resulting in significant savings in terms of memory, computation, and communication resources. Note that FedAvg consumes less memory and wall-clock time because we assign the smallest affordable models to all clients.

Table \ref{tb:communication} lists the total size of data packets transmitted through the network by all clients. We observe that the communication cost of \textit{FedConv}, LotteryFL, and HeteroFL are comparable, as they exclusively transmit sub-model parameters without extra contents. In contrast, Hermes and TailorFL have to transmit the pruning structure, and FedMD needs to transmit logits. Thus, \textit{FedConv}, LotteryFL, and HeteroFL are more friendly to resource-constrained clients. Moreover, it holds significant potential that exiting quantization techniques \cite{abdelmoniem2021towards, dettmers20158} and masking method \cite{DBLP:conf/sensys/0005SZZLC21} can be extended to \textit{FedConv}, to further diminish the communication overhead.

\textbf{Remarks.} In summary, benefiting from the lighter communication and computation burden imposed on resource-constrained clients, \textit{FedConv} saves more system resources and performs inference tasks faster than the baselines.


\vspace{-5pt}
\subsection{Sensitivity Analysis}

\subsubsection{Varying client number.}
We simulate 100 clients and vary the number of selected participating clients from 10 to 50 ($\alpha=10000$) to compare the client model performance with the baselines. As shown in Fig. \ref{fig:varying_client_number_FedConv}, the client model accuracy in \textit{FedConv} exhibits an upward trend as the number of clients increases. For example, the client model accuracy on HARBox increases by 17.54\% when the number of clients increases from 10 to 50. 
We then select CIFAR10 to compare the client model performance of \textit{FedConv} with pruning-based and parameter sharing-based methods. From Fig. \ref{fig:varying_client_number_CIFAR}, we see that \textit{FedConv} attains an average client model accuracy that is at least 32.5\% higher than that of the baselines. The results demonstrate the scalability and superiority of \textit{FedConv} with varying client numbers.

\vspace{-3pt}
\subsubsection{Varying shrinkage ratios.}
To investigate the trade-off between the SR and model performance, we set the SR for 10 clients as 1.0 and set the SR for the remaining 10 clients as $r$. We then vary $r$ from 1.0 to 0.05 and record the average client model accuracy ($\alpha=10000$). From Fig. \ref{fig:varying_SR_FedConv}, we can see that as the SR decreases below a certain threshold, there is a notable accuracy drop in client models, as expected. 
For MNIST, WiAR, and DCD, the SR threshold is about 0.25 (the red shadow), and for CIFAR10, CINIC10, and HARBox, the threshold is about 0.4 (the blue shadow). 
Fortunately, we find that even a lightweight device (\eg, Raspberry Pis) can afford the GoogLeNet model on CINIC10 when the SR is 0.4. Consequently, as long as the SR remains above the corresponding threshold, it can be reduced to conserve system resources effectively. 

We then use CIFAR10 to compare \textit{FedConv} with the baselines, and the client model accuracy is shown in Fig. \ref{fig:varying_SR_CIFAR}. We can see that though the accuracy of \textit{FedConv} also decreases with limited resources, it can retain much higher accuracy than the baselines. 
The reason is that with a lower SR (higher pruning rate), the baselines discard a larger amount of parameter information. In contrast, with \textit{convolutional compression}, \textit{FedConv} can effectively preserve the parameter information of the global model as much as possible to the sub-models bounded by their sizes and resource budgets.

\vspace{-3pt}
\subsubsection{Varying server-side data sizes.} 
To investigate the impact of server-side data, we vary the sample number ratio of the server-side data from 1\% to 25\%, with a step of 0.5\%. As shown in Fig. \ref{fig:varying_server_data}, we obtain two key observations: 1) When the ratio of the server-side data varies from 1\% to 5\%, the client models will have better performance, due to the richer information obtained from the server-side data; 2) After the turning point (> 5\%) the global model tends to overfit to the server-side data, leading to less personalization and degraded client model performance. 
In our evaluation, we set the default sample ratio of the server-side data to 5\%. Note that the actual turning point may differ in practice. In addition, continuous learning \cite{ma2023cost} or incremental learning \cite{he2011incremental} techniques can be further applied as more server-side data become available. 


\begin{figure*}[t]
\vspace{-10pt}
    \centering
    \begin{minipage}[t]{0.49\textwidth}
        \setlength{\abovecaptionskip}{-5pt}
        \subfigtopskip=-5pt
        \subfigcapskip=-5pt
        \centering
        \subfigure[Server-side pre-training]{
            \includegraphics[width=0.46\textwidth]{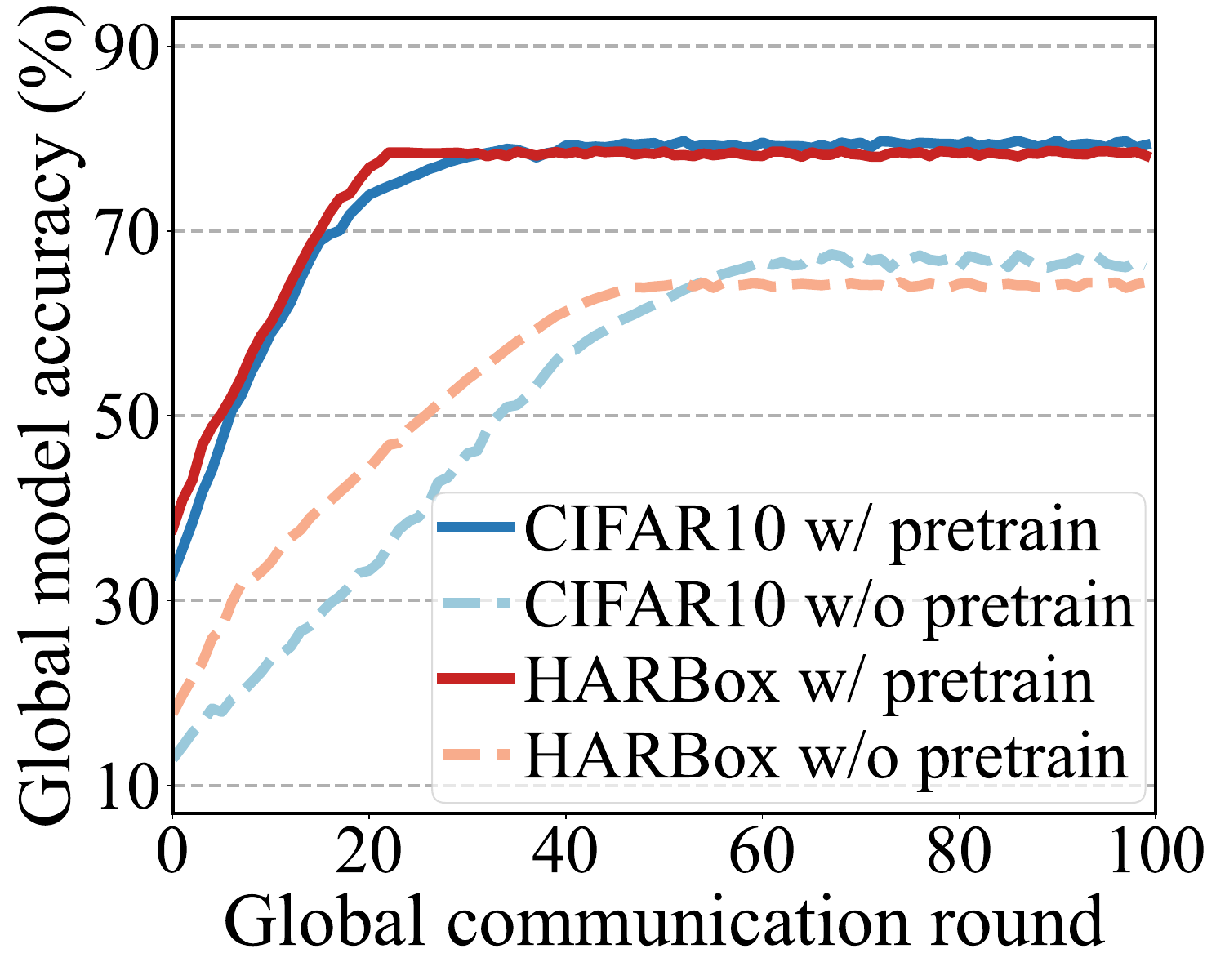}
            \label{fig:ablation_pretrain}
        }
        \centering
        \subfigure[Weight vectors]{
            \includegraphics[width=0.46\textwidth]{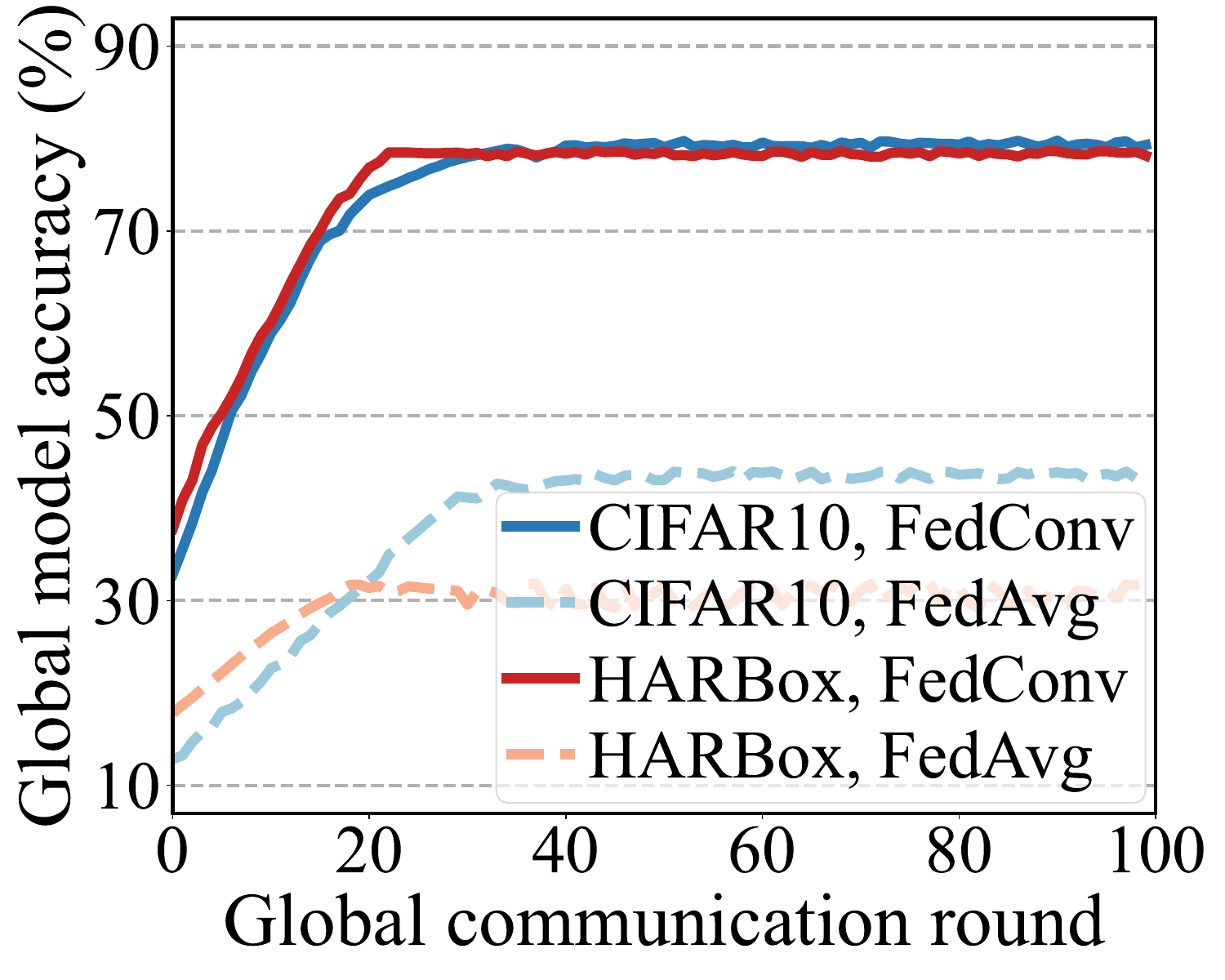}
            \label{fig:ablation_aggregation}
        }
        \caption{Ablation study of \textit{FedConv}.}
    \end{minipage}
    \begin{minipage}[t]{0.49\textwidth}
        \setlength{\abovecaptionskip}{-5pt}
        \subfigtopskip=-5pt
        \subfigcapskip=-5pt
        \centering
        \subfigure[CIFAR10]{
            \includegraphics[width=0.46\textwidth]{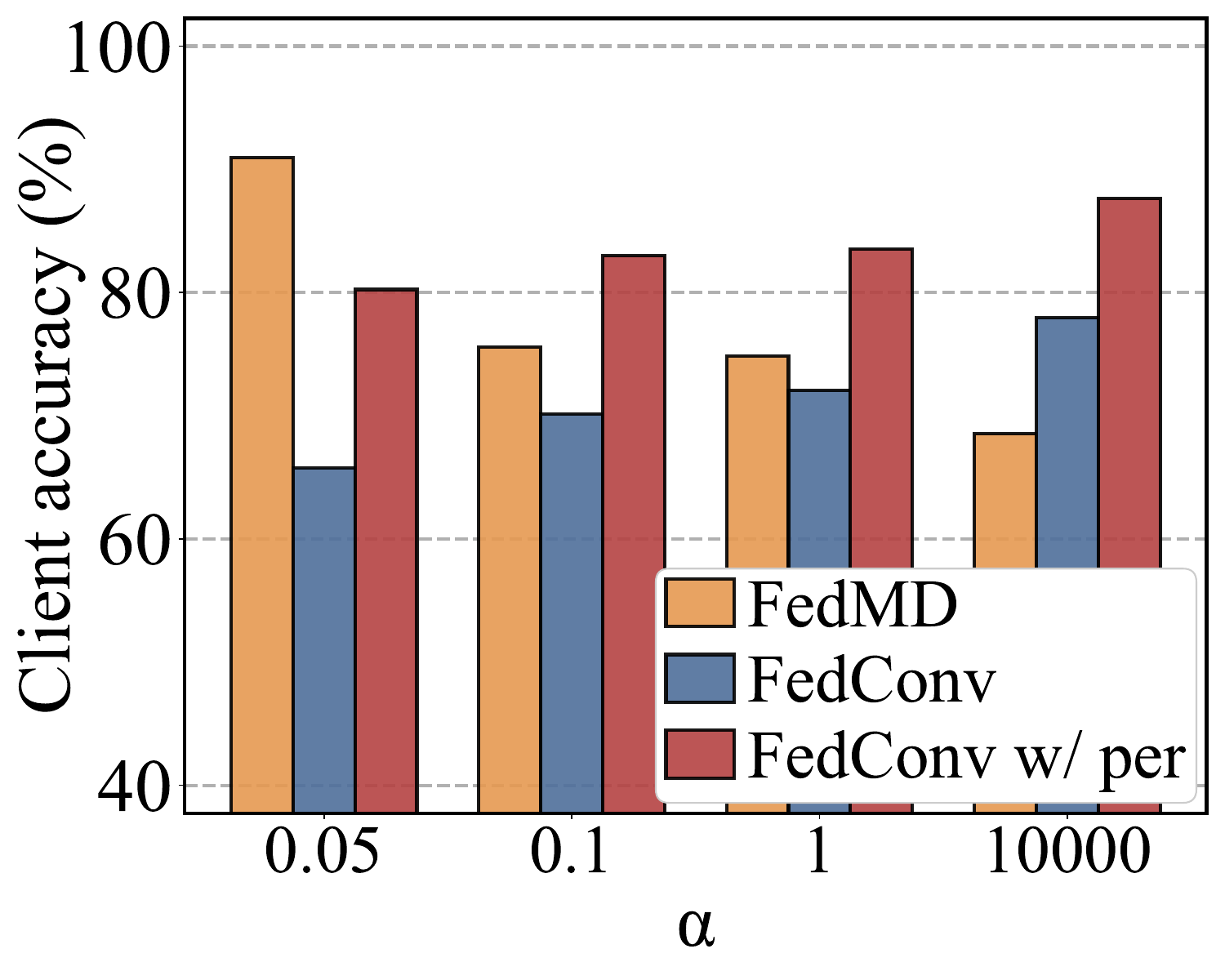}
            \label{fig:personalization_CIFAR10}
        }
        \centering
        \subfigure[HARBox]{
            \includegraphics[width=0.46\textwidth]{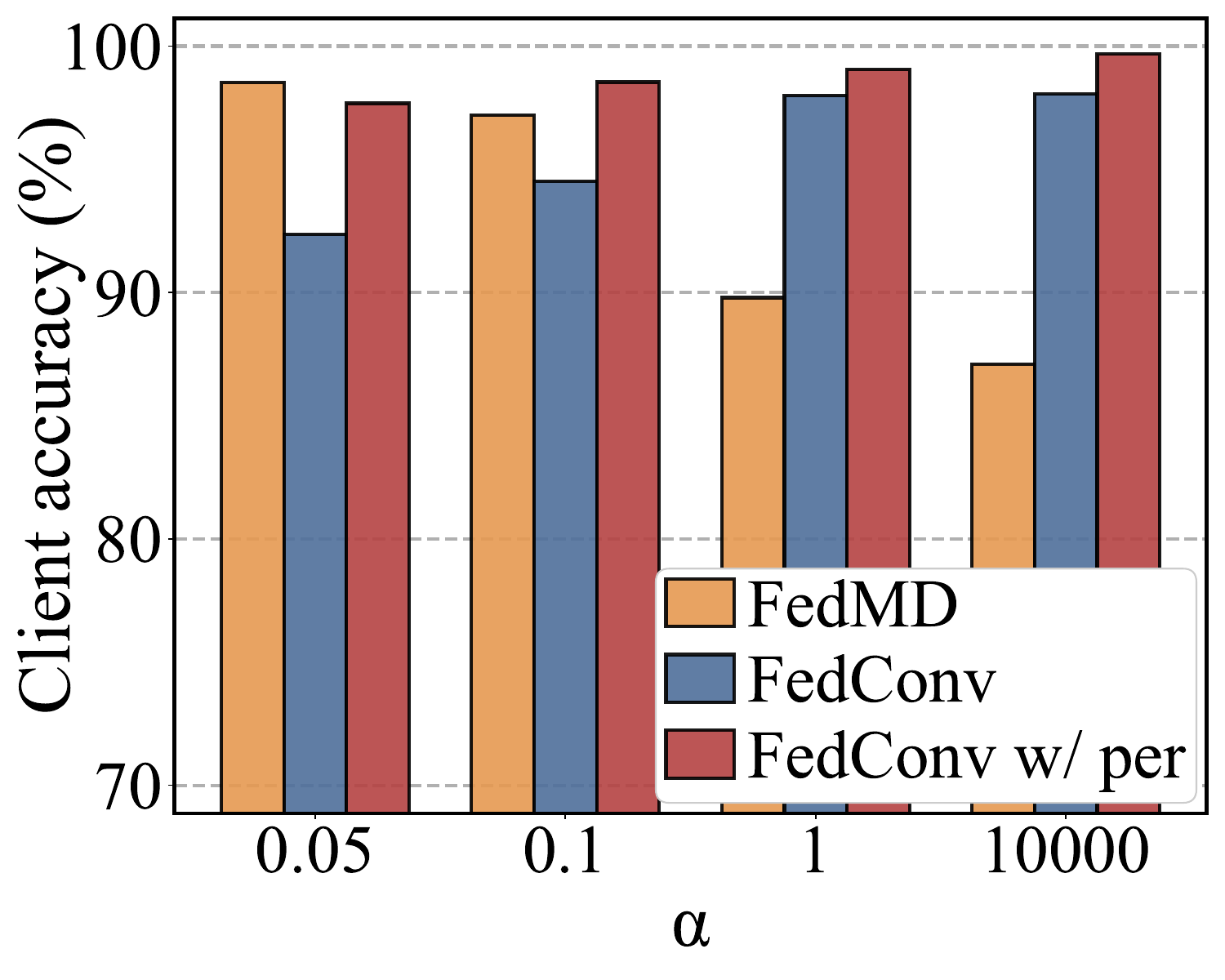}
            \label{fig:personalization_HARBox}
        }
        \caption{\textit{FedConv} with personalized FL.}
        \label{fig:ablation_personalization}
    \end{minipage}
    \vspace{-15pt}
\end{figure*}

\vspace{-3pt}
\subsubsection{Varying hyper-parameters.} 
\label{S:varying_hyper}
We vary the number of epochs for fine-tuning the model compression, dilation, and aggregation to evaluate their impact on the personalization performance of client models. We select two datasets (CIFAR10 and HARBox) for demonstration. We first vary the number of epochs for updating the \textit{convolution/TC parameters} in each global round. Fig. \ref{fig:varying_conv_epoch} shows that when the number of \textit{convolution/TC parameters} updating epochs is around 20, the client models achieve better and more stable performance. After the 20-th and the 40-th epoch, the client model accuracy gradually drops due to the \textit{convolution/TC parameters} being over-fitted to the server-side data. Similarly, from Fig. \ref{fig:varying_aggregation}, we can observe that when the number of tuning epochs for updating \textit{weight vectors} exceeds 40 and 80, the accuracy of the aggregated global model also decreases. Therefore, we set the number of epochs for model compression, dilation, and aggregation to 20.

We also vary the kernel size and stride length of the \textit{compression layers} and report the mean client model accuracy to explore the impact on client performance. We select a convolutional layer from the large model as an example, whose parameter matrix has a shape of $9\times(1,64, 64)$. With the SR being 0.75, the compressed parameter matrix will have a shape of $9\times(1, 48, 48)$. Since the kernel size $k$ and the stride $s$ should satisfy $(64 + 2p - k + 1)/s=48$, the padding $p$ can then be determined accordingly. In general, a larger kernel can capture more comprehensive parameter information, and a smaller stride can capture more fine-grained information. As shown in Fig. \ref{fig:varying_conv_pattern_CIFAR10}, client models tend to have better performance as the kernel size increases and the stride decreases. 
However, a larger kernel incurs high computational complexity and imposes a heavy workload on the server. Therefore, in our default settings, the kernel size and stride are set as 23 and 1, respectively.



\vspace{-5pt}
\subsection{Ablation Study}
Next, we conduct ablation studies to investigate the importance of the server-side pre-training process and the \textit{weighted average aggregation} module, respectively. 

\vspace{-3pt}
\subsubsection{Server-side pre-training.}
Fig. \ref{fig:ablation_pretrain} shows the impact on global model accuracy with and without server-side pre-training with $\alpha=0.05$. It can be observed that with the integration of pre-training, the global model achieves higher average accuracy (about 15.69\%) and reaches faster convergence (about 40 communication rounds earlier), which helps the FL server and clients save communication, computation, and energy costs involved in the training process. 



\vspace{-3pt}
\subsubsection{Weighted average aggregation.}
To demonstrate the impact of our learned \textit{weight vectors} for model aggregation, we assign weights with respect to sample number as in FedAvg to all clients and measure the global model accuracy. Fig. \ref{fig:ablation_aggregation} shows the effect on global model accuracy when performing model aggregation with learned weights and equal weights separately. Significant performance degradation can be observed when employing the averaging aggregation method. This is because the parameters from heterogeneous client models usually exhibit varying skewness toward their local data distribution. Merely averaging all the model parameters overlooks the different contributions made by clients in the aggregation process. On the contrary, with the learned \textit{weight vectors}, clients can contribute different parameter information to the aggregated global model and improve its generalization performance. 


\vspace{-5pt}
\subsection{Personalization Enhancement} 
\label{S: ablation}
To evaluate the potential in personalization, we extend \textit{FedConv} by adding task-specific layers \cite{DBLP:journals/corr/abs-2207-08147} on each client, and evaluate the client model accuracy. Specifically, after receiving the parameters from the server, each client appends its own personal layers to the sub-model. By doing so, the personalization performance of each client can be enhanced during local training. We record the average accuracy of client models after 100 communication rounds. Fig. \ref{fig:ablation_personalization} shows the performance improvement on five datasets after applying personalization enhancement. Compared with FedMD, which achieves the highest client model accuracy (\S~\ref{S:client_model_performance}), we can see that \textit{FedConv} with personalization enhancement is able to surpass FedMD in most cases. This result indicates that \textit{FedConv} can be enhanced with existing personalized federated learning methods to achieve better performance.


\begin{figure}[t]
\vspace{-2pt}
\setlength{\abovecaptionskip}{-3pt}
\subfigcapskip=-7pt
    \centering
    \subfigure[Global model accuracy]{
        \label{fig:case_global}
        \includegraphics[width=0.22\textwidth]{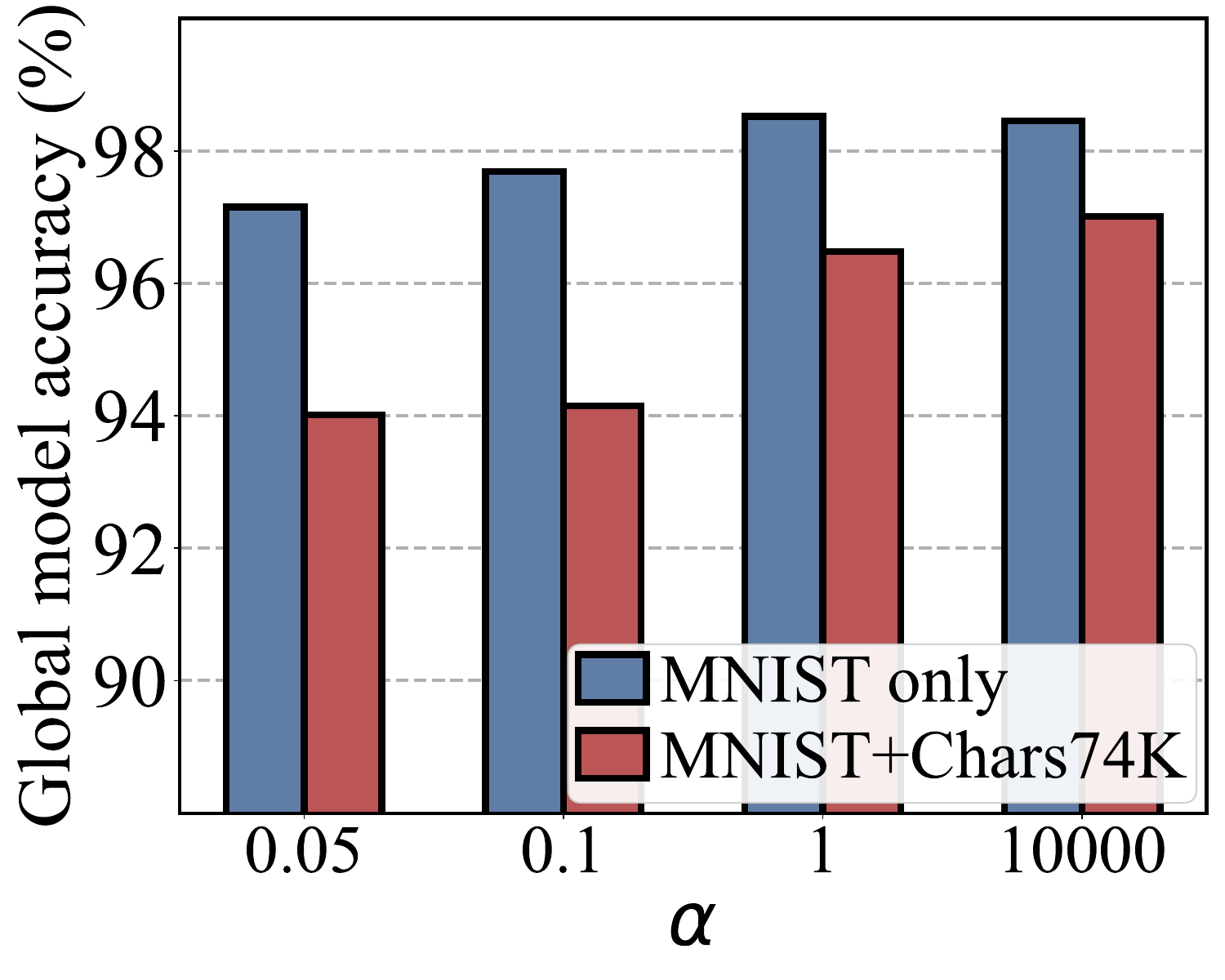}
    }
    \centering
    \subfigure[Client model accuracy]{
        \label{fig:case_client}
        \includegraphics[width=0.22\textwidth]{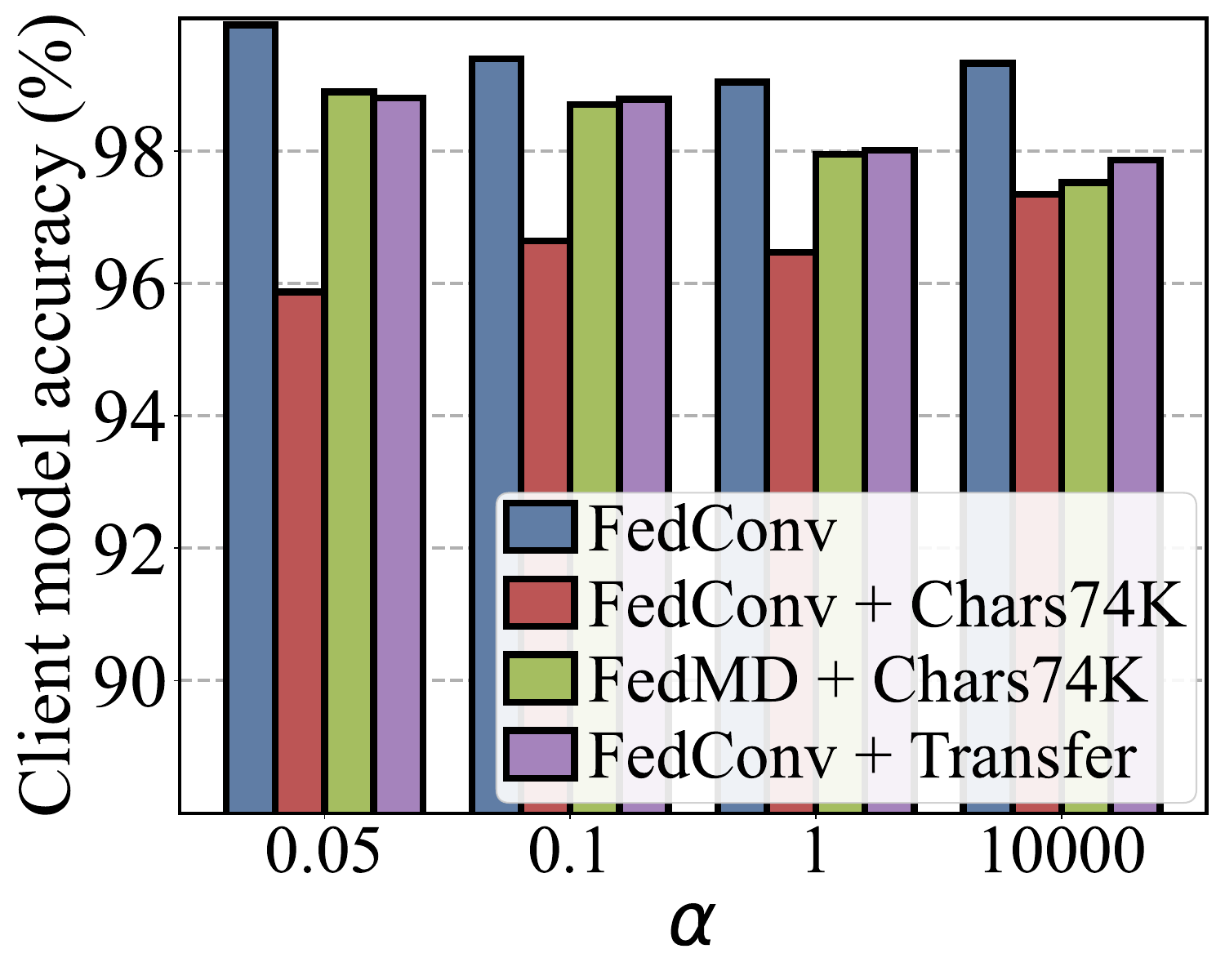}
    }
    \caption{Case study with real-world heterogeneity.}
    \vspace{-15pt}
\end{figure}

\vspace{-5pt}
\subsection{Case Study with Real-World Heterogeneity}
\label{sec:case}
In our default configuration, both the server-side and client-side datasets originate from the same domain. To test with real-world heterogeneity and assess its impact, we conduct a case study where the Chars74K dataset \cite{de2009character} is kept on the server, while the MNIST dataset is used for heterogeneous clients. The Chars74K dataset contains images of digits from computer fonts with variations (italic, bold, and normal). In this case, the global model can learn and extract general features (\eg, different shapes of the digits from the Chars74K dataset), while heterogeneous clients can further fine-tune the compressed model to extract personalized features (\eg, various writing styles of the digits from the MNIST dataset). The \textit{convolutional compression} process and the \textit{TC dilation} process can be regarded as a transformation from one data domain to another. The generated sub-models via \textit{convolutional compression} contain parameter information from the large global model and can thereby extract general features. Similarly, the server applies TC to the locally trained heterogeneous client models to rescale them. This facilitates the aggregation process to form a new global model, retaining the personalization information of the client-side data. As shown in Fig.~\ref{fig:case_global} and Fig.~\ref{fig:case_client}, due to the domain gap between the server-side and the client-side data, there is a decrease in both the global model and the client model accuracy. FedMD still achieves comparable performance to \textit{FedConv} with only the MNIST dataset, benefiting from the knowledge distillation method. However, when we further enhance \textit{FedConv} with transfer learning strategies \cite{saha2021federated, chen2020fedhealth} on each client to narrow down the domain gap between the server-side and the client-side data, the client models will achieve higher accuracies and even outperform FedMD. This observation indicates that \textit{FedConv} can be combined with existing federated transfer learning approaches to achieve better performance.

\vspace{-5pt}
\section{Discussion}

\textbf{Privacy Concerns.} In addition to transferring model parameters between the server and heterogeneous clients, \textit{FedConv} requires all the clients to report their SRs before the FL training starts. To determine appropriate SRs, clients will perform resource profiling locally and report to the server. We note that same as conventional FL schemes, no client-side sensor data needs to be transferred to the server during this process. Thus, we believe the privacy protection of conventional FL schemes can be effectively retained.


\textbf{Practicality of \textit{FedConv}.} In \textit{FedConv}, we use the Flower \cite{DBLP:journals/corr/abs-2007-14390} framework to orchestrate the entire FL process. While Flower offers a stable and robust simulated environment for FL, deploying it in a mixed Linux-Android environment encounters significant obstacles. These include technical challenges in training neural network models on Android devices and issues related to the compatibility of the Flower framework with Android systems. Fortunately, recent advancements \cite{flowerAndroid} in Flower support federated learning setup with Android clients using TensorFlow Lite \cite{abadi2015tensorflow}.

\textbf{\textit{Convolutional Compression}.} As shown in our evaluation, the \textit{convolutional compression} is effective in compressing large global models and achieves better performance compared with model pruning, parameter sharing, and knowledge distillation-based methods. Additionally, the compression and dilation process is performed on the server side, without imposing any extra burden on the clients. From the client's perspective of view, they do not need to participate in the pre-training and fine-tuning process and can join throughout the FL processes, which is the same as the conventional FL systems. 

\vspace{-5pt}
\section{Related Work}

\textbf{Data heterogeneity.} Recent works \cite{DBLP:conf/mobisys/OuyangXZHX21, DBLP:conf/sensys/TuOZHX21, shuai2022balancefl, radovivc2023repa} optimize FL performance under non-IID data. \textit{Clustering-based} methods \cite{DBLP:conf/mobisys/OuyangXZHX21, DBLP:journals/corr/abs-2302-10747, DBLP:conf/cicai/ChenWYHLC22} group clients according to the distribution of their data or model parameters. For example, ClusterFL \cite{DBLP:conf/mobisys/OuyangXZHX21} captures the intrinsic clustering patterns among clients by measuring the similarity of client models. 
Shu \etal \cite{DBLP:journals/iotj/ShuYLCXYJ23} propose a clustered multi-task federated learning on non-IID data. \textit{Personalized FL} adopts local fine-tuning \cite{DBLP:conf/nips/0001MO20} or add task-specific layers on client side \cite{DBLP:conf/nips/YosinskiCBL14, DBLP:journals/corr/abs-2207-08147}. For example, pFedMe \cite{DBLP:conf/nips/DinhTN20} uses Moreau envelopes as a regularized loss function to decouple the task of optimizing a personalized model from the global model learning. Yosinski \etal \cite{DBLP:conf/nips/YosinskiCBL14} enable the upper layers of the global model to learn task-specific features, while the lower layers capture more general features which are further shared across clients. 
Our work is orthogonal to these works and requires minimal modification to clients for integration into existing FL systems.

\textbf{Model heterogeneity and model compression.} 
To accommodate heterogeneous clients, recent works mainly compress the global model to reduce communication and computation costs. They can be divided into three categories: 1) \textit{Knowledge distillation-based methods} \cite{DBLP:conf/nips/ZhangGMWXW21, DBLP:conf/icml/ZhuHZ21} generally regard heterogeneous client models as teacher models and learn an aggregated global student model via knowledge distillation (KD). Lin \etal \cite{DBLP:conf/nips/LinKSJ20} leverage KD and ensemble learning to combine the knowledge from heterogeneous client models. FedMD \cite{DBLP:journals/corr/abs-1910-03581} computes an average consensus to substitute the aggregation process. However, the tuning of KD is performed on clients with a shared dataset, incurring extra overhead for clients; 2) \textit{Parameter sharing strategies} \cite{shen2023feddm} allow sub-models to share a part of the global model parameters to reduce computation overhead. HeteroFL \cite{DBLP:conf/iclr/Diao0T21} enables heterogeneous clients to select fixed subsets of global parameters with minimal modification to the existing FL framework. Yet, the sharing strategy suffers from the imbalance issue \cite{DBLP:journals/corr/abs-2111-14655}; 3) \textit{Pruning-based methods} \cite{vahidian2021personalized} have gained popularity in heterogeneous FL. Hermes \cite{DBLP:conf/mobicom/0005SLPLC21} applies a channel-level pruning method to selectively prune out less important channels. TailorFL \cite{DBLP:conf/sensys/DengCR0LLZ22} proposes an importance value-based filter-level pruning scheme to enable a dual-personalized FL system. Removing entire channels or filters results in information loss and performance degradation \cite{DBLP:conf/iclr/LiuSZHD19}. Unlike these works, we compress the global model with \textit{convolutional compression} to generate sub-models. Orthogonal to our work, traditional compression techniques (\eg, quantization \cite{tonellotto2021neural}) can be applied to compress model parameters and reduce network traffic. However, as the compressed parameters should be decompressed back to their original size before training, these works cannot reduce the system overhead of clients.


\textbf{Convolution and transposed convolution.}
Convolution can effectively extract useful features from input data by capturing local patterns and spatial relationships \cite{DBLP:journals/corr/SimonyanZ14a, DBLP:conf/nips/KrizhevskySH12}. 
In \textit{FedConv}, we exploit a novel \textit{convolutional compression} technique to generate sub-models for heterogeneous clients, which can capture key information embedded in the global model. 
TC is renowned for its capability of reconstructing super-resolution images from fuzzy ones \cite{DBLP:journals/corr/DumoulinV16, DBLP:journals/pami/GaoYWJ20}, which is widely adopted in image dilation \cite{DBLP:conf/iccv/DahlNS17} and semantic segmentation \cite{DBLP:conf/iccv/NohHH15}. In our aggregation process, we leverage TC to resize the heterogeneous client models to a unified size for aggregation.

\vspace{-4pt}
\section{Conclusion}

We propose \textit{FedConv}, a client-friendly federated learning framework for heterogeneous clients, aiming to minimize the system overhead on resource-constrained mobile devices. \textit{FedConv} contributes three key technical modules: 1) a novel model compression scheme that generates heterogeneous sub-models with \textit{convolutional compression} on the global model; 2) a \textit{transposed convolutional dilation} module that converts heterogeneous client models back to large models with a unified size; and 3) a \textit{weighted average aggregation} scheme that fully leverages personalization information of client models to update the global model. Extensive experiments demonstrate that \textit{FedConv} outperforms SOTA baselines in terms of inference accuracy with much lower computation and communication overhead for FL clients. We believe the proposed \textit{learning-on-model} paradigm is worthy of further exploration and can potentially benefit other FL tasks where heterogeneous sub-models can be generated to retain the information of a global model.



\vspace{-4pt}
\begin{acks}
We sincerely thank our shepherd -- Veljko Pejovic, and anonymous reviewers for their constructive comments and invaluable suggestions that helped improve this paper. This work is supported by Hong Kong GRF Grant No. 15206123. This work is also supported by NSFC (Grant No. 62372314, U21A20462, and 62372400), "Pioneer" and "Leading Goose" R\&D Program of Zhejiang under grant No. 2024C03287. Yuanqing Zheng is the corresponding author.
\end{acks}

\vspace{-5pt}
\section*{Appendix}  
  The research artifacts accompanying this paper are available at \href{https://doi.org/10.5281/zenodo.11089994}{https://doi.org/10.5281/zenodo.11089994}.

\clearpage
\balance
\bibliographystyle{ACM-Reference-Format}

\end{sloppypar}
\end{document}